\documentclass[10pt,twocolumn,letterpaper]{article}

\PassOptionsToPackage{table,xcdraw}{xcolor}
\usepackage[pagenumbers]{wacv}      % To produce the REVIEW version for the applications track
\usepackage{graphicx}	
\usepackage{amsmath}	
\usepackage{amssymb}	
\usepackage{booktabs}
\usepackage{times}
\usepackage{microtype}
\usepackage{epsfig}
\usepackage{pifont}
\usepackage{caption}
\usepackage{float}
\usepackage{placeins}
\usepackage{color, colortbl}
\usepackage{stfloats}
\usepackage{enumitem}
\usepackage{tabularx}
\usepackage{xstring}
\usepackage{multirow}
\usepackage{xspace}
\usepackage{url}
\usepackage{subcaption}
\usepackage{bm}
\usepackage[hang,flushmargin]{footmisc}
\usepackage{afterpage}
\usepackage{float}
\usepackage{array}
\usepackage{paralist}

\usepackage{tcolorbox}
\tcbuselibrary{skins}
\newtcolorbox{prompt}[1]{
    enhanced,
    left=4mm,
    right=4mm,
    top=2mm,
    bottom=2mm,
    boxsep=0mm,
    rounded corners,
    title=#1,
    fontupper=\footnotesize\linespread{0.9}\fontfamily{lmr}\selectfont,
    }

% ...

\newcommand{\supp}{Appendix\xspace}

\newcommand{\R}[1]{{%
    \textbf{%
        \ifstrequal{#1}{1}{\textcolor{red}{cSgS}}{%
        \ifstrequal{#1}{2}{\textcolor{blue}{wTs3}}{%
        \ifstrequal{#1}{3}{\textcolor{magenta}{39uS}}{%
        \ifstrequal{#1}{4}{\textcolor{teal}{R#1}}{%
                           \textcolor{cyan}{R#1}%
        }}}}%
    }%
}}

\newcommand{\metric}{\ensuremath{\mathrm{cFreD}}\xspace}

\definecolor{wacvblue}{rgb}{0.21,0.49,0.74}
\usepackage[pagebackref,breaklinks,colorlinks,allcolors=wacvblue]{hyperref}

\def\wacvPaperID{231} % *** Enter the WACV Paper ID here
\def\confName{WACV}
\def\confYear{2026}

\def\paperTitle{Evaluating Text-to-Image and Text-to-Video Synthesis \\ with a Conditional Fr\'echet Distance}

\def\authorBlock{
    Jaywon Koo\thanks{Equal contribution.}\hspace{0.16cm}, Jefferson Hernandez\footnotemark[1]\hspace{0.16cm}, Moayed Haji-Ali, Ziyan Yang, and Vicente Ordonez  \\
    Rice University \\
    {\tt\small \{jk125, jefehern, mh155, zy47, vicenteor\}@rice.edu}  \\
    \small{\url{https://github.com/JaywonKoo17/cFreD}}
}

\title{\paperTitle}
\author{\authorBlock}

\begin{document}
\maketitle
\begin{abstract}
Evaluating text-to-image and text-to-video models is challenging due to a fundamental disconnect: established metrics fail to jointly measure visual quality and semantic alignment with text, leading to a poor correlation with human judgments. 
To address this critical issue, we propose \metric, a general metric based on a Conditional Fr\'echet Distance that unifies the assessment of visual fidelity and text-prompt consistency into a single score. Existing metrics such as %Inception Score (IS), 
Fr\'echet Inception Distance (FID) capture image quality but ignore text conditioning while alignment scores such as CLIPScore are insensitive to visual quality.
Furthermore, learned preference models require constant retraining and are unlikely to generalize to novel architectures or out-of-distribution prompts. 
Through extensive experiments across multiple recently proposed text-to-image models and diverse prompt datasets, \metric exhibits a higher correlation with human judgments compared to statistical metrics
, including metrics trained with human preferences. 
Our findings validate \metric as a robust, future-proof metric for the systematic evaluation of text conditioned models, 
standardizing benchmarking in this rapidly evolving field. We release our evaluation toolkit and benchmark \href{https://github.com/JaywonKoo17/cFreD}{here}.
\end{abstract}  
%\vspace{-0.2in}
\section{Introduction}
\label{sec:intro}
Generative models have demonstrated remarkable capabilities in text-to-image generation~\cite{podell2023sdxl, esser2024scaling, rombach2022high, haji2024elasticdiffusion}, text-to-video generation~\cite{yang2024cogvideox,videoworldsimulators2024,kong2024hunyuanvideo,kling,gen3,seawead}, and other modalities~\cite{haji2024av, brooks2023instructpix2pix, he2021context, mo2024freecontrol, zhang2023adding, haji2024taming}.
This progress has been driven by technical breakthroughs such as the development and improved understanding of Generative Adversarial Networks (GANs)~\cite{NIPS2014_5ca3e9b1}, Variational AutoEncoders (VAEs)~\cite{Kingma2013AutoEncodingVB}, and more recently models based on Denoising Diffusion~\cite{sohl2015deep} and Flow Matching~\cite{lipman2023flow}. 
Automatic evaluation metrics have played a crucial role in guiding the development and refinement of these models by offering quantitative benchmarks that compare the quality, diversity, and fidelity of various generative approaches. 
The most commonly used metrics include Inception Score (IS)~\cite{salimans2016improved}, Fr\'echet Inception Distance (FID)~\cite{heusel2017gans}, and CLIPScore~\cite{hessel2021clipscore}. 

\begin{figure}[t!]
    \centering
    \includegraphics[width=0.98\columnwidth]{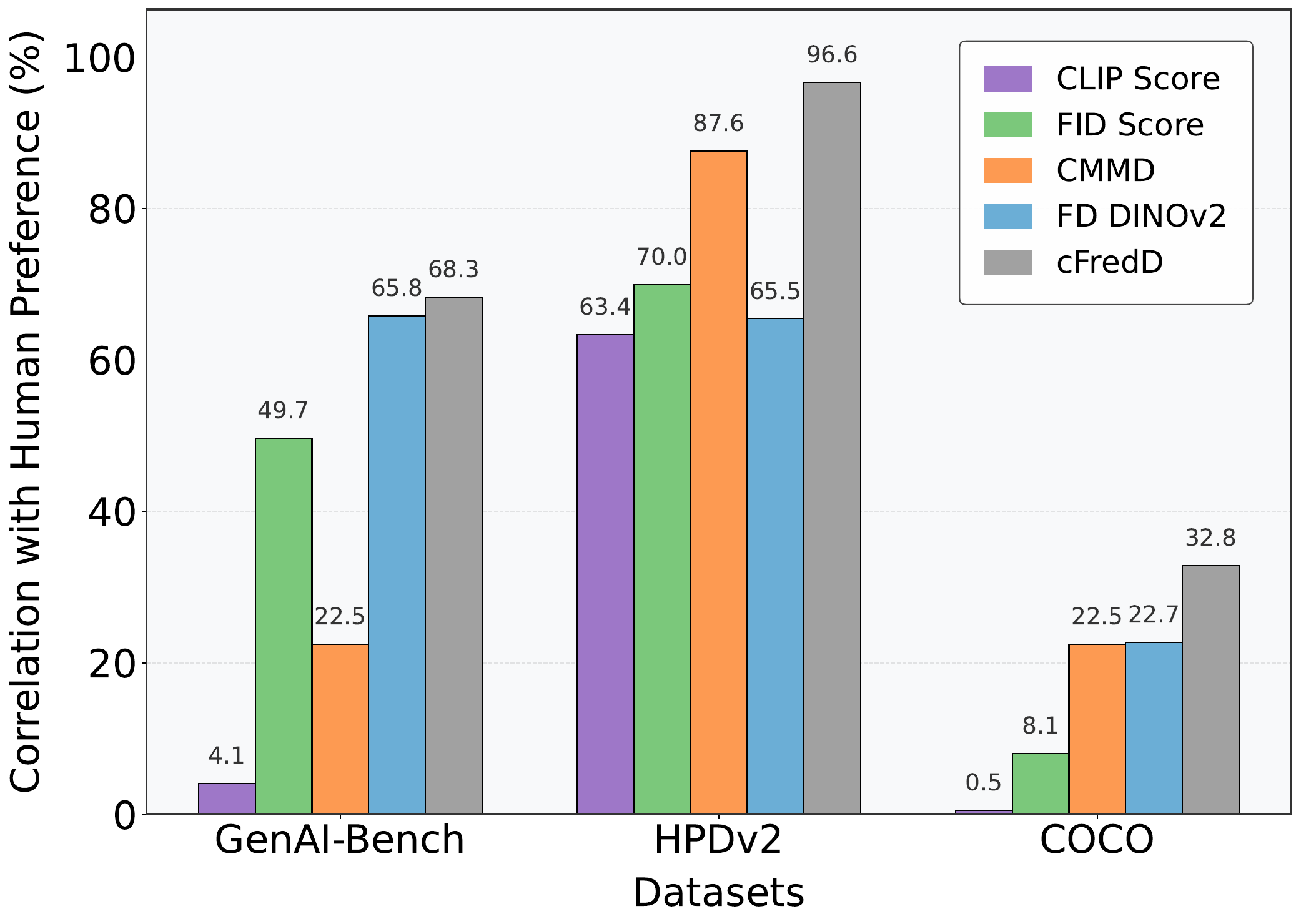}
    \vspace{-0.1in}
    \caption{Correlation of five evaluation metrics with human preferences on three benchmark datasets (GenAI-Bench, HPDv2, and COCO). Compared to FID, $\text{FD}_{\text{\tiny{DINOv2}}}$, CLIPScore, and CMMD, our proposed method (cFreD) achieves consistently higher correlation with human judgments across all datasets.}
    \label{fig:human_correlation}
    \vspace{-0.15in}
\end{figure}

\begin{figure*}[ht!]
    \centering
    \includegraphics[width=0.95\textwidth]{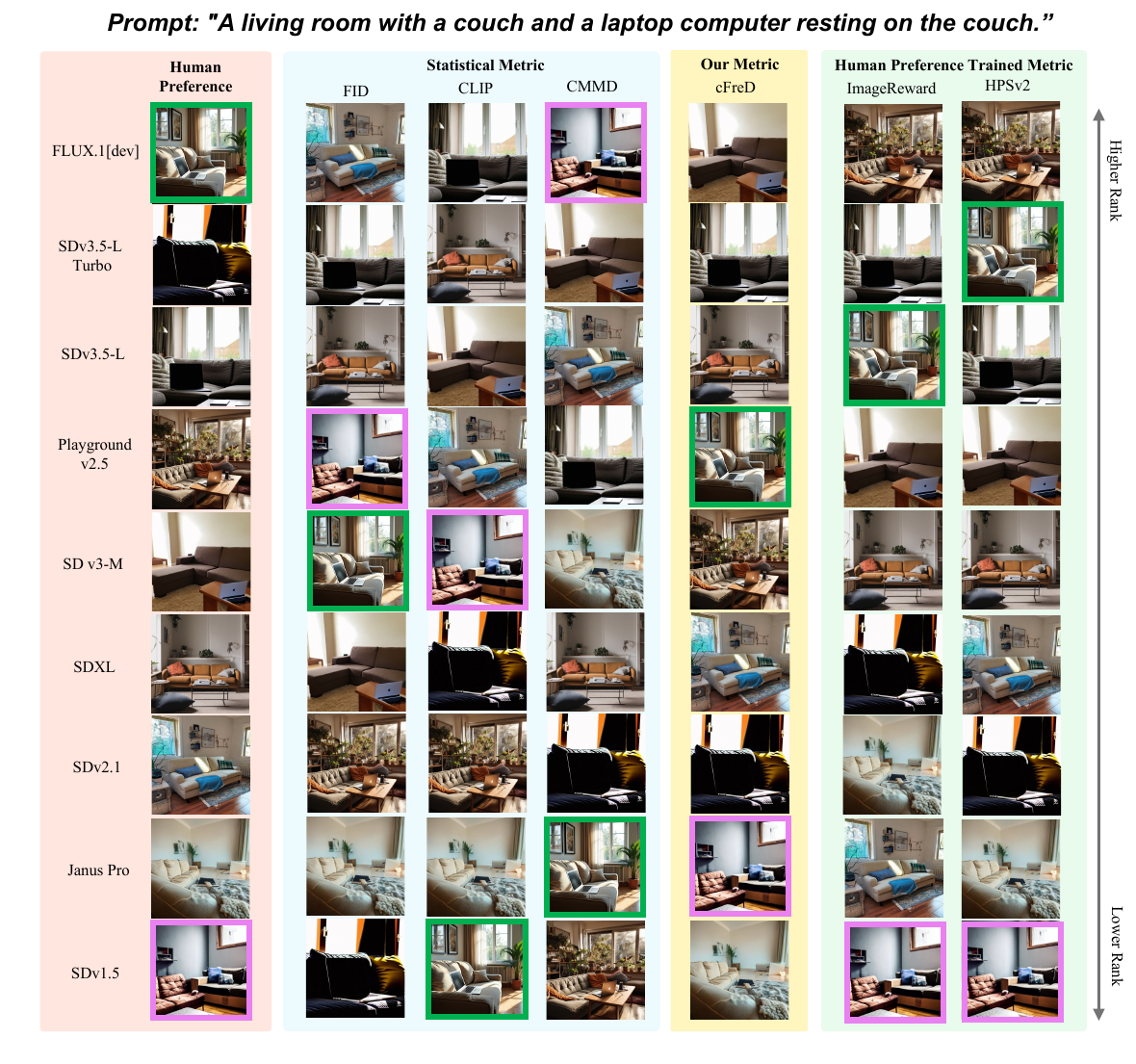}
    \vspace{-0.2in}
    \caption{Comparison of image rankings across different evaluation metrics for the prompt `A living room with a couch and a laptop computer resting on the couch.' Each row displays how images are ranked from 1 to 9 according to a specific metric (Human preference, FID, CLIP, CMMD, ImageReward, and \metric), illustrating how different metrics prioritize distinct visual qualities. Inside a \textbf{\textcolor[rgb]{0.0,0.6,0.0}{green}} box is the highest rated model as judged by human preferences (FLUX.1-dev) and inside a \textbf{\textcolor[rgb]{0.8,0.0,0.8}{purple}} box is the lowest rated model as judged by human preferences (SDv1.5). \metric is the only statistical metric that preserves their relative ordering. }
    \vspace{-0.1in}
    \label{fig:coco}
\end{figure*}

We consider that it is time to reassess evaluation metrics for text-to-image synthesis and propose \metric, based on a conditional formulation of Fr\'echet Distance. 
Prior work has reported that existing metrics exhibit weak correlation with human preferences~\cite{wu2023human, xu2023imagereward, kirstain2023pick, jedi}. 
Additionally, 
metrics such as IS or FID, although indicative of image quality to some extent, do not measure the alignment of generated images with input text prompts, focusing solely on comparing the generated image distributions with respect to a reference set~\cite{BidermanTweet2023}. For instance, consider a dataset with two images: one of a {\em dog} and one of a {\em cat}, each paired with their corresponding prompt. A perfect text-to-image model that mistakenly swaps these mappings (\ie generating a {\em cat} for {\em dog} prompt and vice versa) would achieve near zero FID since the overall distribution of cats and dogs is maintained, despite the misalignment with the intended prompts.
We show that \metric captures better image quality assessment and conditioning on input text and results in improved correlation with human preferences (see~\cref{fig:human_correlation} and~\cref{fig:coco}).
\metric seamlessly extends to text-to-video generation, where it matches the best metric in correlation to human preferences on recent benchmarks, confirming that the same conditional formulation scales to the temporal domain. 
Moreover, we demonstrate that \metric behaves monotonically under severe latent-space image corruptions and prompt perturbations, whereas traditional scores such as FID can be misleading, showcasing its robustness beyond the clean-data regime.

Currently, the most reliable form of evaluation for text-to-image models relies on collecting human preference data through crowd-sourcing, similar to the way Large Language Models (LLMs) are evaluated (i.e.~the LMSys Arena~\cite{chiang2024chatbot}). For instance, the Parti Prompts Arena~\cite{OpenGenAIPartiPromptsLeaderboard} employs 1,600 English prompts originally designed for assessing Parti models~\cite{yu2022scaling}, presenting users with a gamified task of selecting their preferred image from two random outputs generated by four different models played over 10 rounds. Similarly, the Text-to-Image Arena Leaderboard~\cite{ArtificialAnalysisTextToImageArena} asks users to compare the outputs from two randomly selected models given a prompt, and their preferences are used to calculate {\it ELO} scores. However, collecting enough human preference data to statistically establish model superiority is time-intensive, and these arenas risk becoming unmaintained over time, as evidenced by the Parti Prompts Arena, which no longer collects votes. Recent work has also proposed learning metrics directly from human preferences~\cite{xu2023imagereward, kirstain2023pick, wu2023human, wu2023hpsv2, zhang2024learning}. 
However, it remains unclear whether such metrics trained on data from previous generative models can effectively assess the next generation of models. Moreover, since human preferences are inherently dynamic and evolve over time, these models risk becoming outdated unless they are regularly updated with fresh preference data. As a result, metrics such as FID, CLIPScore and our proposed \metric will continue to play an important role in model evaluation.

Most related to our work is the previously proposed conditional Fréchet Inception Distance ($\mathrm{cFID}$) by Soloveitchik et al.~\cite{soloveitchik2021conditional}. Unlike traditional statistical metrics, $\mathrm{cFID}$ incorporates a conditioning modality to evaluate not only visual quality but also alignment with respect to an input condition. We revisit and apply this formulation, originally proposed to evaluate image-to-image generation, to text-to-image generation. Furthermore, we systematically analyze how the choice of vision and text backbone models affect metric performance and demonstrate that the previously used Inception V3~\cite{szegedy2016rethinking} model is an outdated choice. By replacing it with a superior backbone model, we propose \metric as a superior alternative to other statistical metrics (see~\cref{fig:human_correlation}).

Our contributions can be summarized as follows: 

\begin{compactitem}
    \item We demonstrate that a metric based on Conditional Frechet Distance (\metric) is a powerful alternative to the more common practice of evaluating text-to-image models with a combination of FID and CLIPScore. 
    \item We show that \metric correlates strongly with human evaluations, outperforming prior statistical metrics on four text-to-image human preference datasets and matching the best metric on two text-to-video human preference datasets.
    \item We open-source our evaluation toolkit and offer the community a unified version of the Partiprompts Arena.
\end{compactitem}

\section{Related Work}
\label{sec:related}

\noindent\textbf{Text-to-Image Generation Metrics.}
Text-to-image synthesis has progressed from early GANs~\cite{zhang2017stackgan,xu2018attngan} to today’s diffusion and autoregressive models such as GLIDE, Imagen, and DALLE~\cite{nichol2021glide,saharia2022photorealistic,ramesh2021zero}.  
Quality is still commonly measured with Fréchet Inception Distance (FID) and Inception Score (IS)~\cite{heusel2017gans,salimans2016improved}, which emphasize image realism/diversity but correlate only loosely with human judgments and ignore text alignment~\cite{jayasumana2024rethinking}.  
To address this, CLIP-based metrics, including CLIPScore~\cite{hessel2021clipscore}, CMMD~\cite{jayasumana2024rethinking}, and $\text{FD}_{\text{\footnotesize DINOv2}}$~\cite{stein2023exposing}, compare image–text pairs in a joint representation space, while alternatives such as FWD~\cite{veeramacheneni2023fr} (wavelet statistics) and perceptual LPIPS~\cite{zhang2018unreasonable} target visual fidelity alone. Despite these refinements, existing scores still under-represent the nuanced trade-off between image quality and semantic faithfulness that guides human preference.  Our proposed \metric integrates the text prompt directly into the distance computation, producing evaluations that align more closely with human ratings of text-to-image outputs.

\noindent\textbf{Text-to-Video Generation Metrics.} 
Early work repurposed image-centric scores: Fréchet Video Distance (FVD) on I3D features captures visual realism but misses motion/text cues, whereas frame-wise CLIPScore~\cite{hessel2021clipscore} adds a text component yet remains motion blind. 
T2VBench introduces 16 controlled temporal scenarios that expose consistency failures, while EvalCrafter~\cite{liu2024evalcrafter} uses 17 objective metrics to measure perceptual quality, motion, and text alignment.
JEDi~\cite{jedi} replaces FVD’s Gaussian assumption with an MMD on JEPA embeddings, reducing sample needs
, while learning-based evaluators such as VideoScore~\cite{he2024videoscore} align well with human judgments but demand expensive
retraining, and don't measure the alignment of the generative distribution beyond human preferences.
In contrast, our \metric simultaneously captures temporal and perceptual fidelity \emph{and} text alignment without retraining.

\noindent \textbf{Evaluating with Human Preferences.}
A complementary line of research replaces statistical metrics with direct human feedback, assembling large‐scale preference datasets in which annotators pick or rank images for a given prompt~\cite{kirstain2023pick,xu2023imagereward,wu2023human,wu2023hpsv2,zhang2024learning,MPS,peng2024dreambench++,fu2023dreamsim,ku2023viescore}.  
Early efforts such as HPS~\cite{wu2023human} and PickScore~\cite{kirstain2023pick} crowd-sourced $\sim$100k and $\sim$500k pairwise votes, respectively, then fine-tuned CLIP (L/14 or H/14) 
so that ``preferred'' images obtain higher image–text similarity.  
ImageReward~\cite{xu2023imagereward} (137k pairs), HPSv2~\cite{wu2023hpsv2} (108k GPT-cleaned prompts), and MPS~\cite{MPS} (66k multi-aspect rankings) further extend scale, data sanitation, and evaluation axes; training reward models that outperform FID/IS on human correlation.
While these datasets improve alignment with user preferences, they demand expensive annotation and still entangle aesthetics, realism, and semantic faithfulness.  
Our work pursues a lighter-weight alternative: a metric that leverages textual context without relying on human‐preference supervision.
Unlike preference-trained metrics, which may overfit to current model families seen during training and struggle with small synthetic (e.g., CIFAR-100) or domain-specific data (e.g., aerial or medical imaging), \metric requires no extra training and offers a reliable plug-and-play solution across diverse settings. 
\section{Methodology}
\label{sec:method}
In this work, we evaluate text-to-image synthesis using a conditional variant of the Fr\'echet Distance (FD). This formulation was also proposed in the work of Soloveitchik et al.~\cite{soloveitchik2021conditional} as a conditional version of the FID metric. For succinctness, we offer here a review of this formulation. Unlike the standard Fr\'echet Distance (FD) which only compares the marginal distributions of real and generated images, \emph{conditional FD} accounts for the conditioning provided by a text prompt $x$. 
\begin{table*}[t]
\centering
%\footnotesize
%\setlength{\tabcolsep}{1pt}
\renewcommand{\arraystretch}{1.1}
\scalebox{0.9}{
\begin{tabular}{lccccccccccc}
\toprule
\multirow{2}{*}{\textbf{Dataset}}&
\multicolumn{2}{c}{\textbf{FID}}&%$\downarrow$} & 
\multicolumn{2}{c}{$\text{\textbf{FD}}_{\text{{\textbf{DINOv2}}}} $} & %\downarrow$} & 
\multicolumn{2}{c}{\textbf{CLIP} } &   %$\uparrow$} &  
\multicolumn{2}{c}{\textbf{CMMD} } &  %$\downarrow$} & 
\multicolumn{2}{c}{\textbf{cFreD}} %$\downarrow$}
 \\
\cmidrule(lr){2-3} \cmidrule(lr){4-5} \cmidrule(lr){6-7} \cmidrule(lr){8-9} \cmidrule(lr){10-11}
   & $\rho^2$ &  Rank Acc. & $\rho^2$ & Rank Acc.  & $\rho^2$ & Rank Acc. & $\rho^2$ & Rank Acc. & $\rho^2$ & Rank Acc. \\
\midrule
\textbf{HPDv2~\cite{wu2023hpsv2}}      & 0.70  & 86.7 & 0.65  & 86.7 &  0.63   &  15.6   &  0.88 & 80.0 &\textbf{0.97} & \textbf{91.1}            \\
\textbf{Gen-AI Bench~\cite{lin2024evaluating}} &  0.50 &  80.0 &  0.66  &  86.7 & 0.04  &  40.0    &  0.22 & 66.7 & \textbf{0.68} & \textbf{86.7} \\
\textbf{PartiPrompts~\cite{OpenGenAIPartiPromptsLeaderboard}} &  0.70 & - & 0.70 & - & 0.12 & -  & 0.54  & -  &  \textbf{0.73} & -               \\
\textbf{COCO~\cite{lin2014microsoft}} &   0.08 & 41.7 & 0.29 & 36.1  & 0.00  & 47.2  & 0.22 & 30.6 &\textbf{0.33} & \textbf{66.7}            \\
\midrule
\textbf{Avg.}   &  0.50 &  69.5 &  0.58 & 69.8 & 0.20 & 34.3 & 0.47 & 59.1 & \textbf{0.68} & \textbf{81.5}            \\
\bottomrule
\end{tabular}
}
%\vspace{-1em}
\caption{Correlation to human preference and rank accuracy of statistical metrics on four text-to-image datasets (HPDv2, Parti-Prompts, COCO and Gen-AI Bench). The best results are shown in \textbf{bold}. \metric consistently achieves the highest correlation with human preference and rank accuracy across all four datasets.
}
\label{tab:text_to_img_overall}
\vspace{-1em}
\end{table*}

\subsection{Conditional Fr\'echet Distance}
For a given prompt $x$, we assume that the real image $y$ and the generated image $\hat{y}$ are drawn from the conditional distributions
\begin{equation}
    \begin{aligned}
    Q_{y|x} &= \mathcal{N}(\mu_{y|x},\, \Sigma_{yy|x}), \\
    Q_{\hat{y}|x} &= \mathcal{N}(\mu_{\hat{y}|x},\, \Sigma_{\hat{y}\hat{y}|x}),
    \end{aligned}
\end{equation}
where $\mu_{y|x}$ and $\mu_{\hat{y}|x}$ are the conditional means, and $\Sigma_{yy|x}$ and $\Sigma_{\hat{y}\hat{y}|x}$ are the conditional covariance matrices. These statistics capture the distribution of images corresponding to the text prompt.

Following the derivation in ~\cite{soloveitchik2021conditional}, the conditional Fr\'echet Distance is defined as the expectation over the prompts of the distance between the two conditional Gaussian distributions:

\begin{equation}
    \label{eq:cfid_1}
    \begin{aligned}
    \metric &= \mathbb{E}_{x}\Bigg[\|\mu_{y|x} - \mu_{\hat{y}|x}\|^2 \nonumber\\[1mm]
    & \quad +\, \mathrm{Tr}\Big(\Sigma_{yy|x} + \Sigma_{\hat{y}\hat{y}|x} - 2\Big(\Sigma_{yy|x}^{1/2} \Sigma_{\hat{y}\hat{y}|x} \Sigma_{yy|x}^{1/2}\Big)^{1/2}\Big)\Bigg].
    \end{aligned}
\end{equation}

An equivalent formulation can be derived that expresses \metric in terms of the unconditional moments and the cross-covariances:
\begin{equation}
    \label{eq:cfid}
    \begin{aligned}
    \metric = & \|\mu_y - \mu_{\hat{y}}\|^2  + \mathrm{Tr}\Big[(\Sigma_{yx} - \Sigma_{\hat{y}x}) \Sigma_{xx}^{-1} (\Sigma_{xy} - \Sigma_{x\hat{y}})\Big] \nonumber\\[1mm]
    &+\, \mathrm{Tr}\Big[\Sigma_{yy|x} + \Sigma_{y\hat{y}|x} - 2\Big(\Sigma_{yy|x}^{1/2} \Sigma_{\hat{y}\hat{y}|x} \Sigma_{yy|x}^{1/2}\Big)^{1/2}\Big].
    \end{aligned}
\end{equation}
In these expressions, $\mu_y$ and $\mu_{\hat{y}}$ are the unconditional means of the real and generated images, respectively. In contrast, $\Sigma_{yx}$ and $\Sigma_{x\hat{y}}$ denote the cross-covariances with the input prompt $x$.

By incorporating the prompt $x$, \metric simultaneously measures (1) the realism of the generated images (via the comparison of unconditional statistics) and (2) the fidelity of the generated images with respect to the text prompt (via the conditional moments and cross-covariances).
Thus, \metric provides a more comprehensive evaluation metric for text-to-image models compared to traditional FID, which may fail to penalize cases where a model produces realistic images that are uncorrelated with the input prompt.

\section{Experiment Settings}
\label{sec:experiments}
In this section, we introduce the experimental setup and report the results on two tasks. We next compare it with preference-trained metrics, variants of \metric using different backbones, and the FID $+$ CLIPScore combination.
\begin{table*}[t]
\centering
%\footnotesize
%\setlength{\tabcolsep}{1pt}
\renewcommand{\arraystretch}{1.1}
\scalebox{0.9}{
\begin{tabular}{lccccccccccc}
\toprule
\multirow{2}{*}{\textbf{Dataset}}&
\multicolumn{2}{c}{\textbf{FVD}}&%$\downarrow$} & 
\multicolumn{2}{c}{\textbf{JEDi}} & %\downarrow$} & 
\multicolumn{2}{c}{\textbf{IS} } &   %$\uparrow$} &  
\multicolumn{2}{c}{\textbf{CLIP} } &  %$\downarrow$} & 
\multicolumn{2}{c}{\textbf{cFreD}} %$\downarrow$}
 \\
\cmidrule(lr){2-3} \cmidrule(lr){4-5} \cmidrule(lr){6-7} \cmidrule(lr){8-9} \cmidrule(lr){10-11}
   & $\rho^2$ &  Rank Acc. & $\rho^2$ & Rank Acc.  & $\rho^2$ & Rank Acc. & $\rho^2$ & Rank Acc. & $\rho^2$ & Rank Acc. \\
\midrule
\textbf{T2VQA-DB~\cite{kou2024subjective}}      & 0.82  & 86.7 & 0.69  & 82.2 &  0.59   &  73.3  &  0.01 & 57.8 &\textbf{0.86} & \textbf{91.1}  \\
%\textbf{T2VQA-DB~\cite{kou2024subjective}}      & 0.818  & 86.67 & 0.690  & 82.22 &  0.589   &  73.33  &  0.012 & 57.78 &\textbf{0.860} & \textbf{91.11}  \\
\textbf{EvalCrafter~\cite{liu2024evalcrafter}} &  \textbf{0.93} &  80.0 &  0.60  &  20.0 & 0.79  &  80.0    &  0.46 & 40.0 & 0.90 & \textbf{100} \\
%\textbf{EvalCrafter~\cite{liu2024evalcrafter}} &  \textbf{0.928} &  80.00 &  0.601  &  20.00 & 0.788  &  80.00    &  0.460 & 40.00 & 0.899 & \textbf{100.0} \\
\midrule
\textbf{Avg.}   &  \textbf{0.88} &  83.35 &  0.65 & 51.1 & 0.69 & 76.7 & 0.24 & 48.9 & \textbf{0.88} & \textbf{95.6} \\
%\textbf{Avg.}   &  0.873 &  83.34 &  0.65 & 51.11 & 0.69 & 76.67 & 0.24 & 48.89 & \textbf{0.880} & \textbf{95.56}            \\
\bottomrule
\end{tabular}
}
%\vspace{-1em}
\caption{Correlation to human preference and rank accuracy of statistical metrics on two text-to-video datasets (T2VQA-DB and EvalCrafter). The best results are shown in \textbf{bold}. \metric achieves the highest average correlation and rank accuracy with two datasets, demonstrating strong alignment with human judgement in text-to-video evaluation.
}
\label{tab:text_to_vid_overall}
\vspace{-1em}
\end{table*}
\subsection{Experimental Setup}
\label{subsec:Experimental_Setup}

\vspace{0.02in}
\noindent \textbf{Dataset.} 
To evaluate how correlated our metric is with human preferences, we need rankings of generated images from different models using the same prompt. 
We use three different datasets (HPDv2~\cite{wu2023hpsv2}, Gen-AI Bench~\cite{lin2024evaluating}, and PartiPrompts~\cite{OpenGenAIPartiPromptsLeaderboard}), which contain different generated images (each from a different model) for each prompt, along with the human preferences.  
We took the most preferred image per prompt as the reference image, since the real image was not always rated highest by humans. 
To evaluate recent models, we randomly selected 1,000 prompts from the COCO train and validation sets, ensuring these prompts were not part of the HPDv2 training and testing sets. For each prompt, we generated images using nine models listed on the Arena Leaderboard~\cite{ArtificialAnalysisTextToImageArena}. The original COCO images serve as reference images in this evaluation.

\vspace{0.02in}
\noindent \textbf{Metric Comparison \& Evaluation Setting. }
We compare \metric against four statistical metrics (FID~\cite{heusel2017gans}, $\text{FD}_{\text{\footnotesize{DINOv2}}}$~\cite{stein2023exposing}, CLIPScore~\cite{hessel2021clipscore}, and CMMD~\cite{jayasumana2024rethinking}). 
For text-to-video, we also compare with four statistical metrics (FVD~\cite{unterthiner2019fvd}, JEDi~\cite{jedi}, Inception Score, and CLIPScore).
We evaluate human correlation $\rho$ for scoring perspectives and calculate the Rank Accuracy to compare ranks. 
\metric uses \texttt{DINOv2-G/14}~\footnote{\href{https://huggingface.co/timm/vit_large_patch14_dinov2.lvd142m}{timm/vit\_giant\_patch14\_dinov2.lvd142m}} for image embedding and the \texttt{OpenCLIP ConvNext-B Text Encoder}~\footnote{ \href{https://huggingface.co/laion/CLIP-convnext_base_w-laion2B-s13B-b82K-augreg}{convnext\_base\_w.laion2b\_s13b\_b82k\_augreg}} for text embeddings. For text-to-video, \metric uses \texttt{} for text embedding and FVD's \texttt{I3D} for video embeddings.

\vspace{0.03in}
\noindent \textbf{Rank Accuracy.} 
This metric quantifies the proportion of correctly ordered pairs—that is, it represents the probability that a randomly selected pair is concordant with the true rank.
Previous works on learning human preferences~\cite{schuhmann2022laion, xu2023imagereward, wu2023hpsv2, MPS} measure performance using per item rank accuracy, which computes the ranking accuracy for each image-text pair and then averages the results. In contrast, \metric evaluates the overall ranking performance across the entire dataset by computing a global rank accuracy. For statistical metrics~\cite{heusel2017gans, stein2023exposing, hessel2021clipscore, jayasumana2024rethinking}, we follow this convention and derive rankings directly from their raw scores. For human preference-trained metrics, we first average the ranking each model receives over all samples and then determine the final ranking based on these averages.

\subsection{Results on Text-to-Image}
\label{subsec:Results_text-to-image}
% \subsubsection{Human Preference Dataset v2}
As shown in Table~\ref{tab:text_to_img_overall}, \metric achieves the highest correlation with human preference and the highest rank accuracy among all statistical metrics across the four benchmarks.
FID and $\text{FD}_{\text{\footnotesize{DINOv2}}}$ show only moderate alignment 
, while the CLIP score performs poorly and CMMD provides only marginal improvements. In contrast, \metric consistently outperforms all baselines, achieving the highest correlation and ranking accuracy on every dataset.
We provide full per-dataset breakdowns in the supplementary material. 
These results suggest that \metric provides a significantly more reliable and generalizable metric for human perceptual evaluation in text-to-image generation tasks. 

\begin{figure*}[t]
    \centering
    % First row: three images
    \begin{subfigure}[b]{0.48\textwidth}
        \centering
        \includegraphics[width=\textwidth]{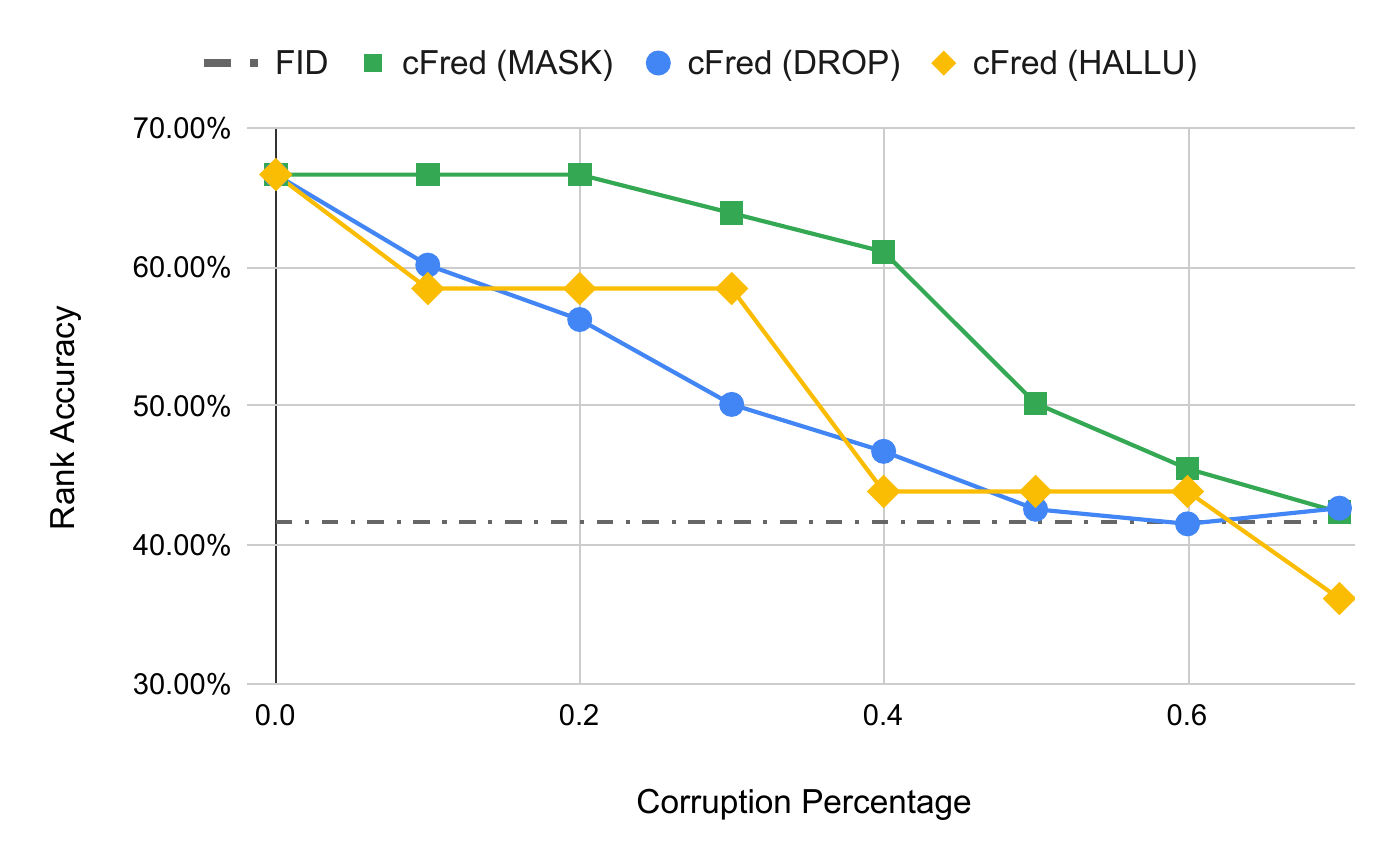}
        \caption{\textbf{Robustness of text distortions.} We progressively corrupt the caption embeddings by (i) replacing random tokens with \texttt{[MASK]} (MASK),  (ii) randomly dropping tokens (DROP), or (iii) hallucinating new tokens with an external LLM (HALLU).}
        \label{fig:subfig_txt_distortions}
    \end{subfigure}
    \hfill
    \begin{subfigure}[b]{0.48\textwidth}
        \centering
        \includegraphics[width=\textwidth]{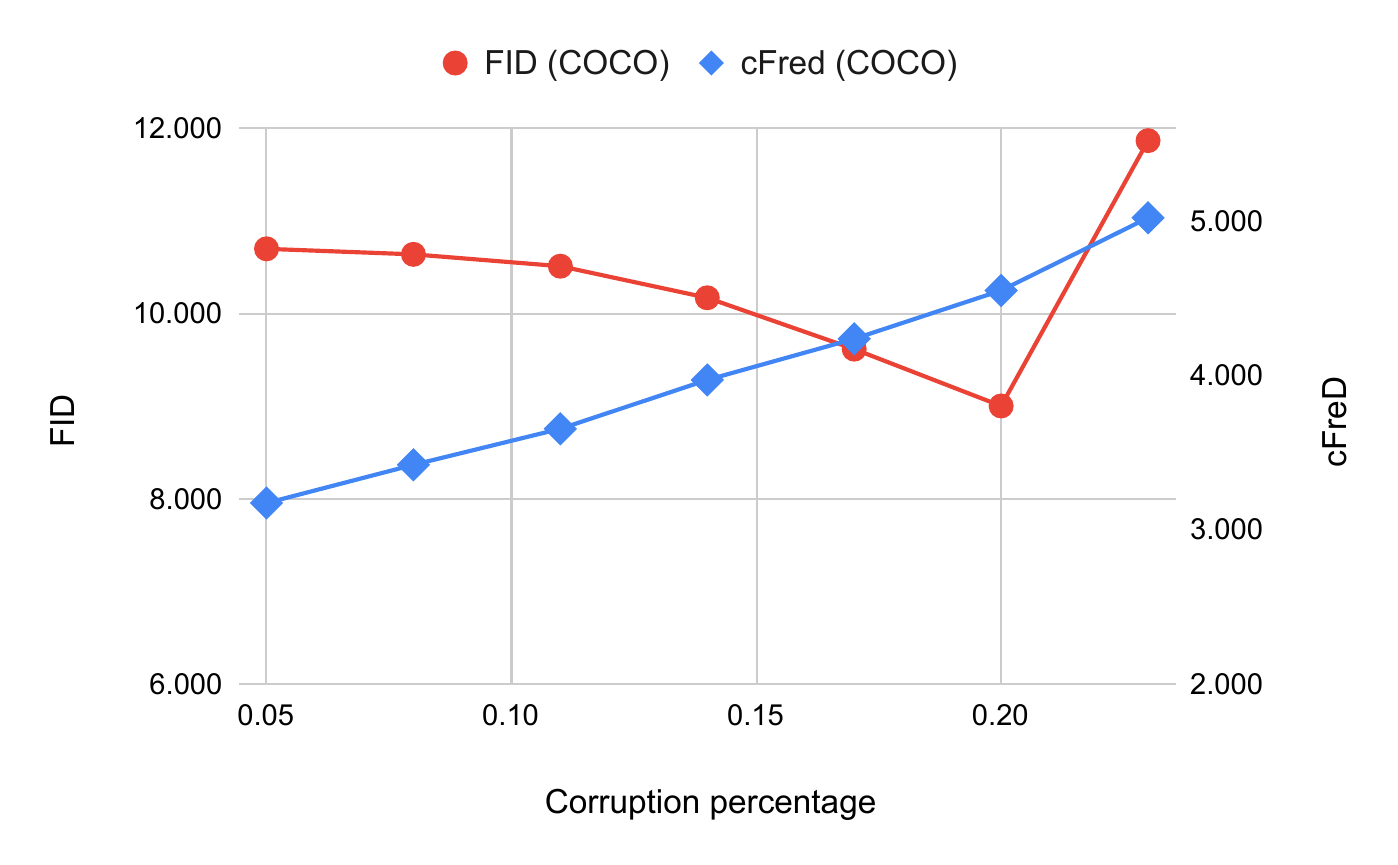}
        \caption{\textbf{Robustness of image distortions.} Images are first encoded with a VQ-VAE; we then progressively add uniform noise to a growing fraction of latent codes before decoding.  Here \metric\ rises monotonically with corruption, faithfully tracking visual degradation.}
        \label{fig:subfig_img_distortions}
    \end{subfigure}

    \caption{\textbf{Robustness study on the COCO.} dataset We inject noise either into the \emph{text} branch \textbf{(a)} or the \emph{image} branch \textbf{(b)} and compare \metric\ with the conventional FID baseline.   Across both settings, \metric\ responds consistently to quality degradation, whereas FID can be insensitive (\textbf{a}) or misleading (\textbf{b}).}
    \label{fig:distortions_ablation}
    \vspace{-0.1in}
\end{figure*}

\cref{fig:coco} shows how different evaluation metrics (FID, CLIP, CMMD, ImageReward, and \metric) rank the same set of images from best to worst for the prompt, with human preferences highlighting the best (green box) and worst (purple box) outputs. Notably, \metric is the only statistical metric that preserves the human-chosen ordering. 

\begin{figure}[t!]
    \centering
    \includegraphics[width=\columnwidth]{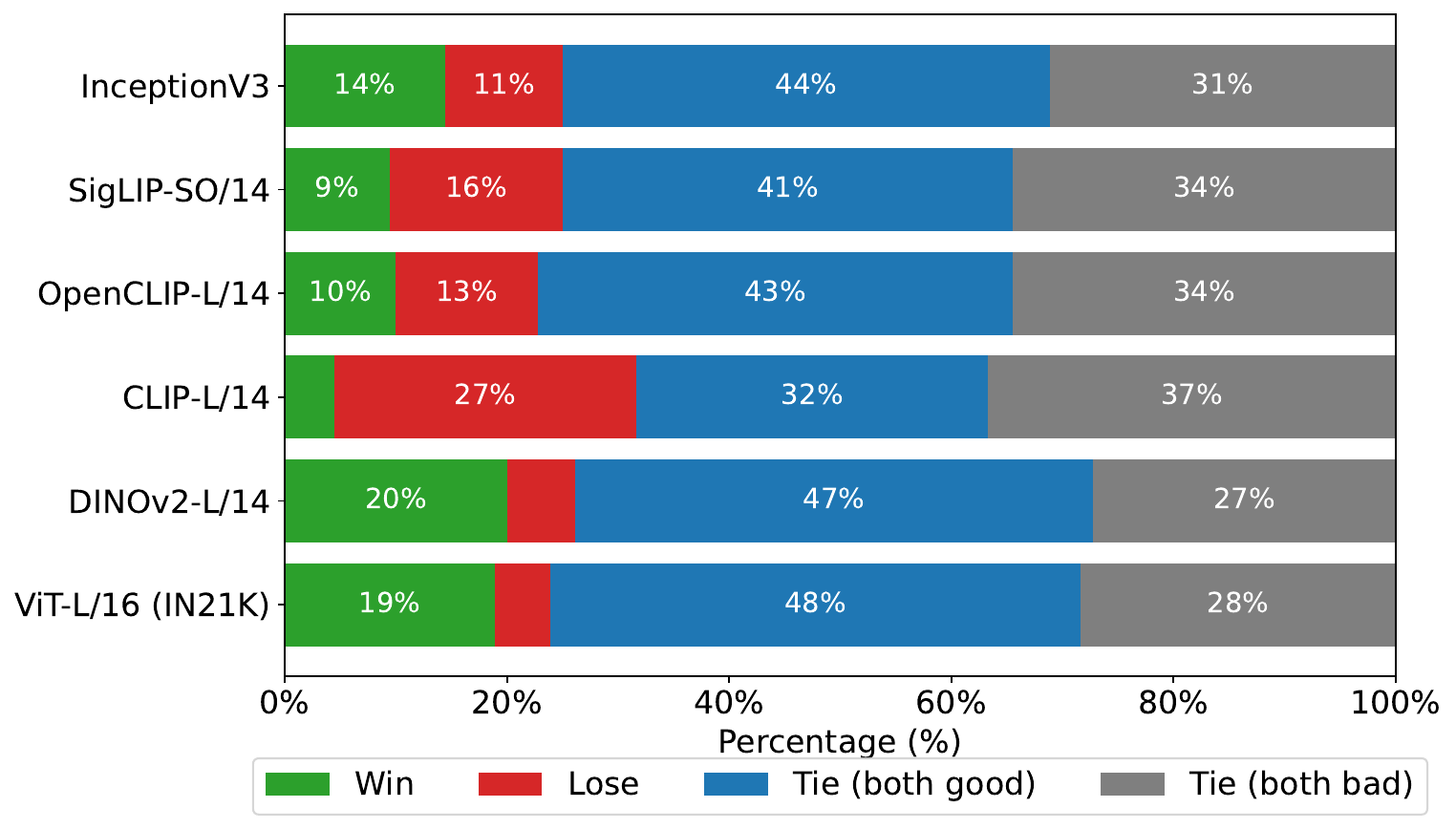}
    \caption{\textbf{Win rates} between the concordance of the true rank against the ranking created by each image backbone when evaluated on the COCO dataset.}
    \label{fig:backbone_comparison}
\vspace{-0.15in}
\end{figure}

\subsection{Results on Text-to-Video}
\label{subsec:Results_text-to-video}

\cref{tab:text_to_vid_overall} shows the comparison of \metric with four other text-to-video statistical metrics across two benchmarks: T2VQA-DB~\cite{kou2024subjective} and EvalCrafter~\cite{liu2024evalcrafter}. \metric consistently achieves the highest rank accuracy and among the highest correlations with human preference across both datasets. While FVD performs strongly in terms of correlation (avg. $\rho^2 = 0.88$), its rank accuracy is notably lower at 83.35\%. Other baselines, such as JEDi, IS, and CLIP, perform worse. These results highlight the robustness and generalizability of \metric, demonstrating its closer alignment with human judgment in text-to-video evaluation.

\subsection{Are we done with FID?}
\label{subsec:are_we_done_fid}
FID and CLIPScore have been the main metrics for evaluating text-to-image synthesis. We performed an extra experiment where we explored a linear combination of these two metrics optimized on the COCO dataset. Although this combined metric achieved a non-trivial rank accuracy of 60\%, its performance still lagged behind that of \metric. More critically, when applied to the HPDv2 dataset, the combined metric's rank accuracy lowered to 30\%—worse than using FID or CLIPScore individually. 

These results not only question the utility of combining FID and CLIPScore, but also highlight the inherent limitations of FID in capturing the text–image alignment necessary for the evaluation of text-to-image generation. Based on our experiments on PartiPrompts, HPDv2, and COCO, we argue that researchers should shift their focus towards monitoring \metric and CLIPScore as the primary evaluation metrics.

\subsection{Are we done with InceptionV3?}
\label{subsec:are_we_done_inceptionv3}

For this experiment, we include InceptionV3 as one of the backbones for \metric. Our experimental results, summarized in ~\cref{fig:backbone_comparison}, show a head-to-head comparison of each image backbone’s ranking of the diffusion models against the ground-truth ranking, reported as a ``win rate''. Specifically, for each backbone, we measure how often its ranking of the diffusion models agrees with the actual human-derived ordering (win), disagrees (lose), or ties (split into both good or both bad). Although InceptionV3 still secures a portion of wins, it lags behind newer transformer-based backbones, such as DINOv2-L/14 and ViT-L/16, which consistently achieve higher win rates and fewer mismatches. These findings suggest that, although InceptionV3 remains a good baseline, more advanced architectures yield rankings that better align with human judgment on the COCO dataset, reinforcing the move away from InceptionV3 in state-of-the-art text-to-image evaluations.

\begin{table}[t!]
\centering
\setlength{\tabcolsep}{3pt}
\renewcommand{\arraystretch}{1.15}

\scalebox{0.62}{
\begin{tabular}{lccccccccccc}
\toprule
\multicolumn{9}{c}{\textbf{(a) Text-to-Image Evaluation}} \\
\midrule
\multirow{2}{*}{\textbf{Dataset}}&
\multicolumn{2}{c}{\textbf{HPSv2}} &   
\multicolumn{2}{c}{\textbf{MPS}} & 
\multicolumn{2}{c}{\textbf{VQAScore}} &
\multicolumn{2}{c}{\textbf{cFreD}} \\
\cmidrule(lr){2-3} \cmidrule(lr){4-5} \cmidrule(lr){6-7} \cmidrule(lr){8-9}
   & $\rho^2$ & Rank Acc. & $\rho^2$ & Rank Acc. & $\rho^2$ & Rank Acc.  & $\rho^2$ & Rank Acc. \\ 
\midrule
\textbf{HPDv2~\cite{wu2023human}}      &  0.90   & 88.9   &  0.86 & 86.7 & 0.62 & 80.0 & \textbf{0.97} & \textbf{91.1} \\
\textbf{Gen-AI Bench~\cite{lin2024evaluating}} &  0.79 & 87.7 & 0.45 & 73.3 & \textbf{0.99} & \textbf{93.3} & 0.68 & 86.7 \\
\bottomrule
\end{tabular}
}

\vspace{0.5em}

\setlength{\tabcolsep}{6pt}

\scalebox{0.7}{
\begin{tabular}{lcccccc}
\toprule
\multicolumn{7}{c}{\textbf{(b) Text-to-Video Evaluation}} \\
\midrule
\multirow{2}{*}{\textbf{Dataset}}&
\multicolumn{2}{c}{\textbf{VideoScore 1.1}}& 
\multicolumn{2}{c}{\textbf{VisionReward}} & 
\multicolumn{2}{c}{\textbf{cFreD}} \\
\cmidrule(lr){2-3} \cmidrule(lr){4-5} \cmidrule(lr){6-7}
   & $\rho^2$ & Rank Acc. & $\rho^2$ & Rank Acc. & $\rho^2$ & Rank Acc. \\
\midrule
\textbf{T2VQA-DB~\cite{kou2024subjective}} & 0.02 & 55.6 & 0.01 & 46.67 & \textbf{0.86} & \textbf{91.1} \\
\textbf{EvalCrafter~\cite{liu2024evalcrafter}} & \textbf{0.91} & 20.0 & 0.27 & 60.0 & 0.90 & \textbf{100.0} \\
\bottomrule
\end{tabular}
}

\caption{Comparison of \metric (cFreD) with metrics trained using human preference across (a) text-to-image and (b) text-to-video evaluation settings. We report squared Spearman correlation ($\rho^2$) and rank accuracy (\%) with human preference. Best results are shown in \textbf{bold}.}
\label{tab:human_preference_trained_metrics}
\vspace{-0.15in}
\end{table}

\subsection{Comparison with Human Preference Trained Metrics}
\label{subsec:compare_human_preference_trained_metrics}

\cref{tab:human_preference_trained_metrics} compares \metric with existing metrics trained on human preference across text-to-image and text-to-video evaluation settings. In the text-to-image domain (Table~\ref{tab:human_preference_trained_metrics}a), \metric consistently achieves the highest correlation and rank accuracy on HPDv2 and remains competitive on Gen-AI Bench, despite not using any human preference-based training. In the text-to-video domain (Table~\ref{tab:human_preference_trained_metrics}b), \metric outperforms all preference-trained metrics on T2VQA-DB and achieves perfect rank accuracy on EvalCrafter, despite not using any human preference training. Notably, while VideoScore~\cite{he2024videoscore} achieves a slightly higher correlation on EvalCrafter ($\rho^2=0.91$), its rank accuracy drops to 20.0\%, underscoring a gap between correlation and true preference alignment. These results highlight the generality and robustness of \metric across two tasks. Unlike metrics trained on preference data, \metric requires no re-training and can be readily applied to specialized domains such as medical or satellite images without any human annotation or domain-specific adaptation.

\section{Ablations}
To disentangle how image and text encoders influence \metric, we exhaustively evaluated all combinations of vision and language backbones (full list in the \supp).  
Ideally, \emph{A/B} experiments require training identical architectures while varying a \emph{single} factor to isolate causal effects most cleanly.  However, re-training a model on a billion-scale image corpus for every factor would demand too many resources. Rather than reporting raw correlations bucketed by individual factors, we approximate a controlled “one-factor-at-a-time’’ experiment by fitting (i) a multiple linear regression (R\(^2\!=\!0.56\); \(N\!=\!1804\)) and (ii) a mixed-effects model with random intercepts for each image encoder.  These models provide \emph{partial-dependence} estimates that isolate the marginal impact of every visual attribute on \metric’s Spearman correlation with human preferences while averaging over all text encoders.  All analyses are performed on the \textsc{PartiPrompts} Arena.  Comprehensive statistical outputs—including Type-II ANOVA tables, variance-inflation factors, and diagnostic plots—are included in the \supp. 

\begin{figure*}[t]
    \centering
    % First row: three images
    \begin{subfigure}[b]{0.32\textwidth}
        \centering
        \includegraphics[width=\textwidth]{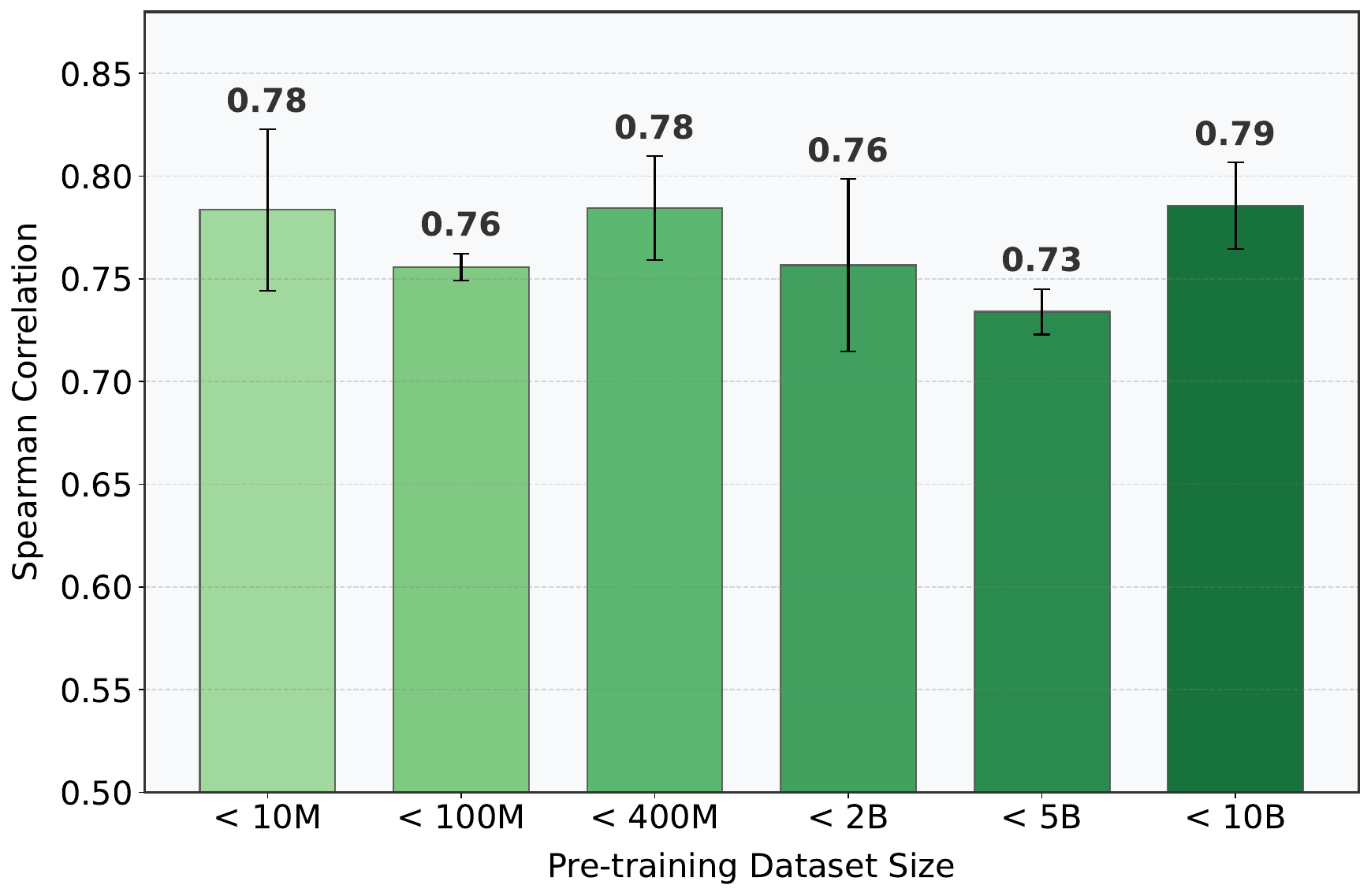}
        \caption{Effect of training data.}
        \label{fig:subfig_data_size}
    \end{subfigure}
    \hfill
    \begin{subfigure}[b]{0.32\textwidth}
        \centering
        \includegraphics[width=\textwidth]{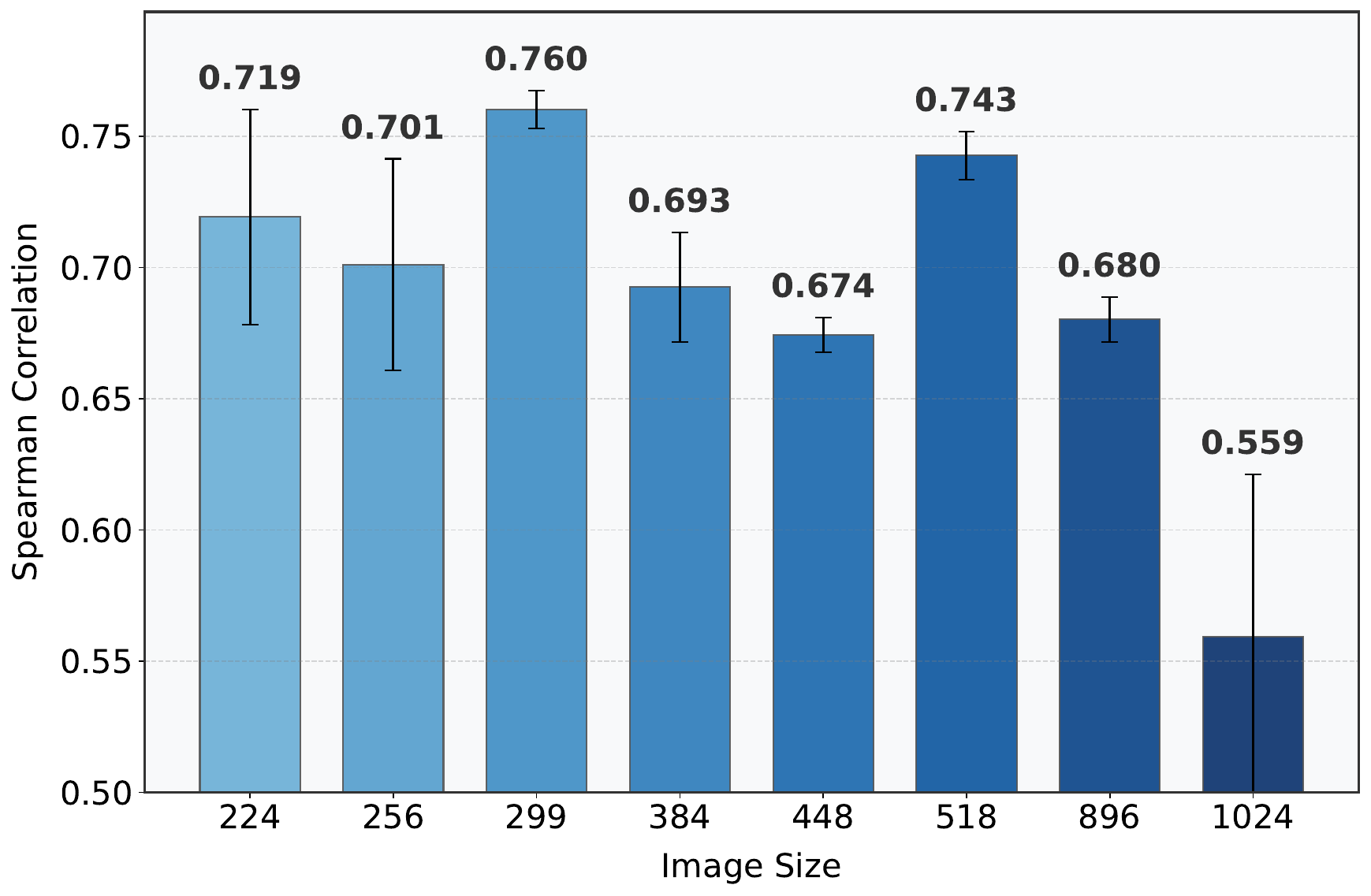}
        \caption{Effect of image size.}
        \label{fig:subfig_img_size}
    \end{subfigure}
    \hfill
    \begin{subfigure}[b]{0.32\textwidth}
        \centering
        \includegraphics[width=\textwidth]{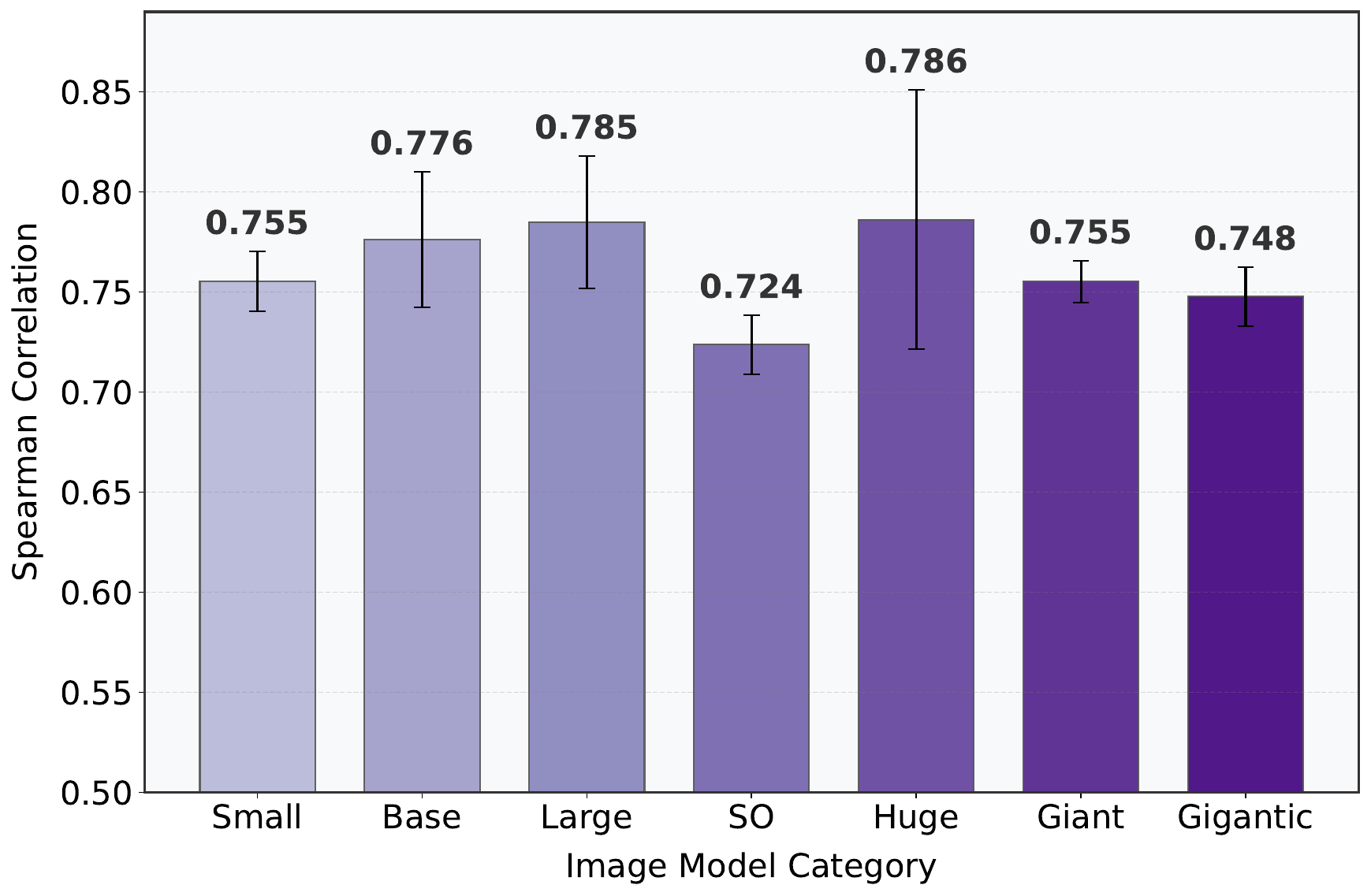}
        \caption{Effect of model size.}
        \label{fig:subfig_model_size}
    \end{subfigure}
    
    \vspace{1.em}
    
    % Second row: two images centered under the three above
    \begin{minipage}{0.7\textwidth}
        \centering
        \begin{subfigure}[b]{0.46\textwidth}
            \centering
            \includegraphics[width=\textwidth]{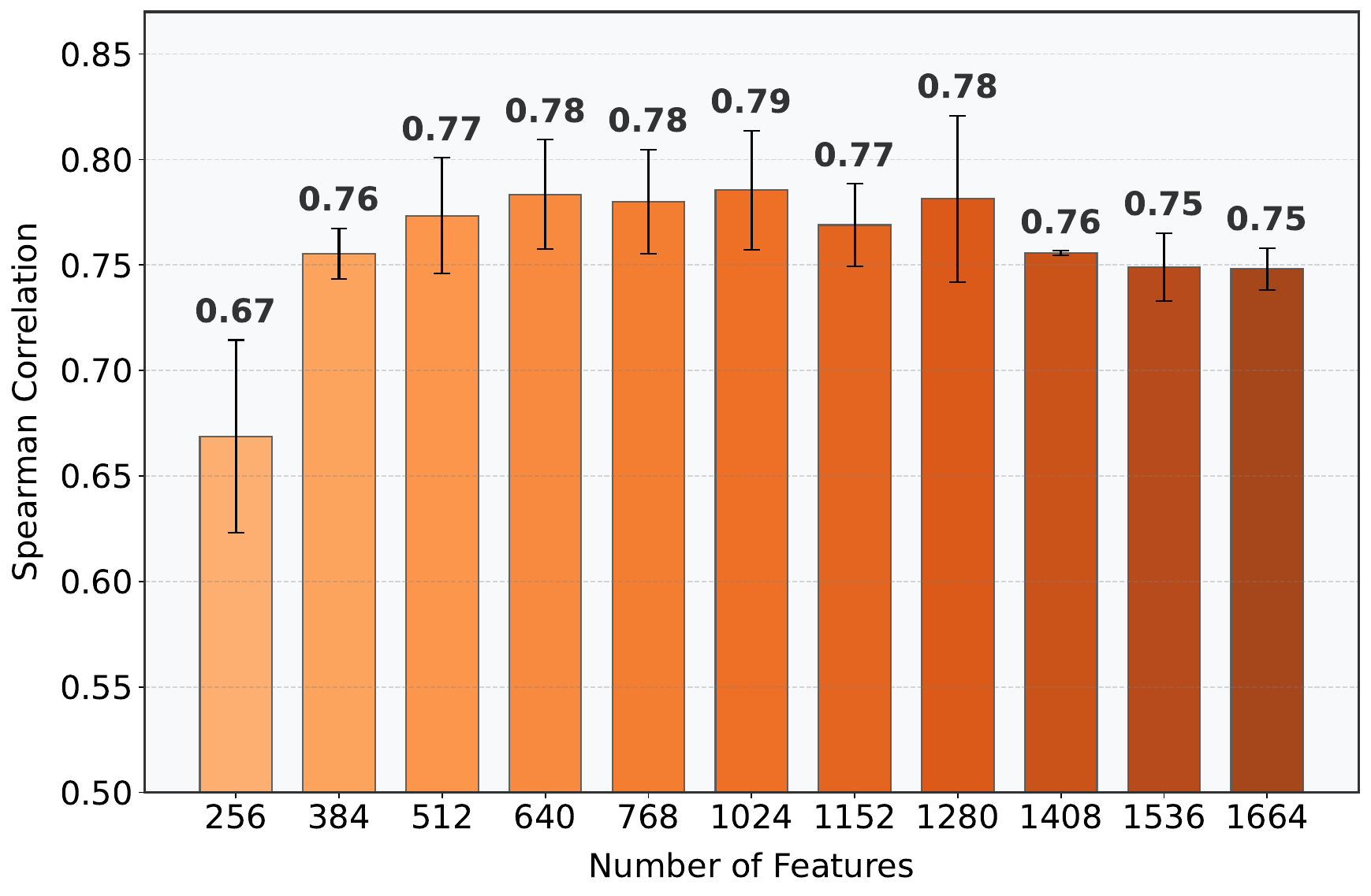}
            \caption{Effect of number of features.}
            \label{fig:subfig_num_features}
        \end{subfigure}\hfill
        \begin{subfigure}[b]{0.46\textwidth}
            \centering
            \includegraphics[width=\textwidth]{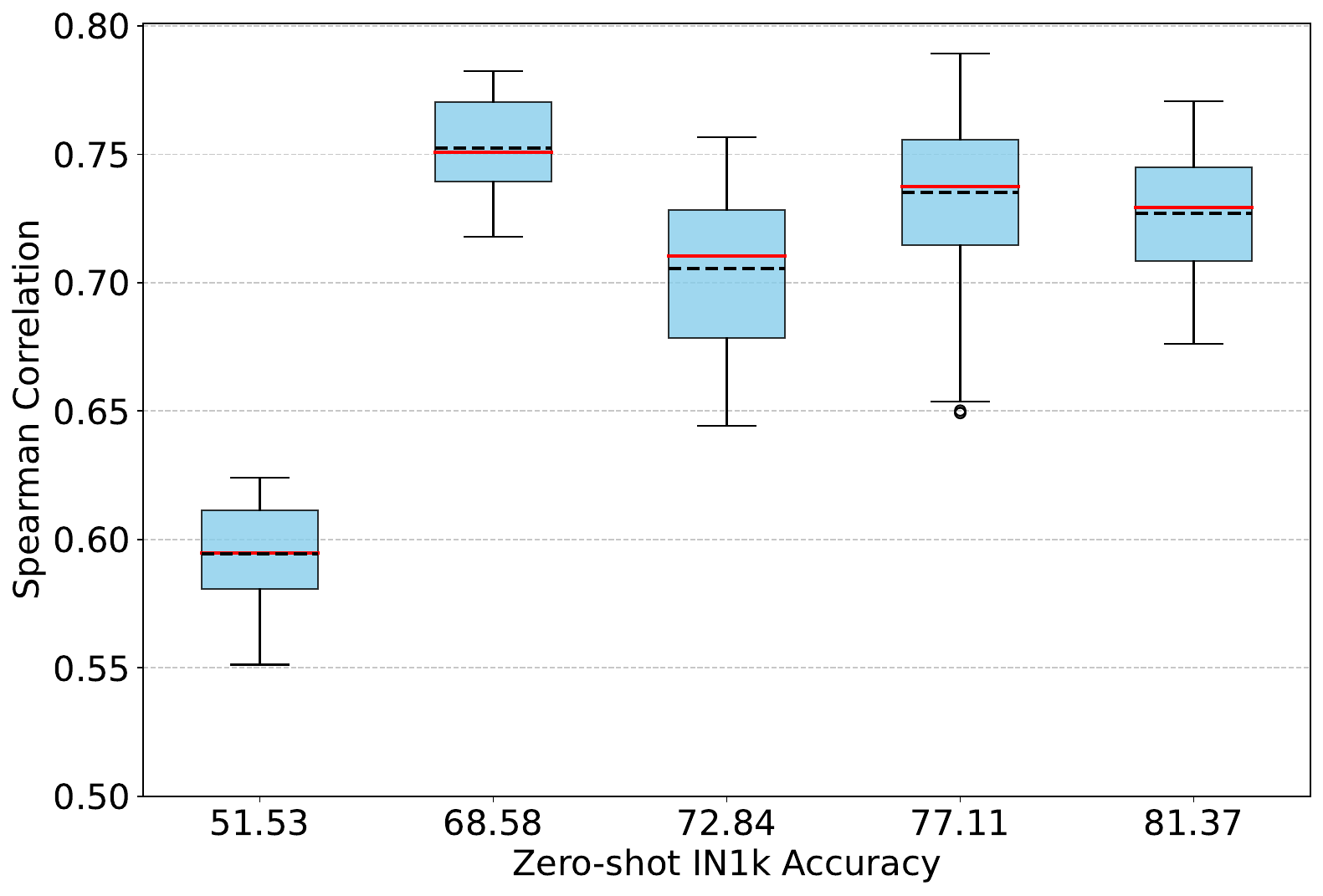}
            \caption{Effect of IN1K Zero-shot accuracy.}
            \label{fig:subfig_zs_in1k}
        \end{subfigure}
    \end{minipage}
    
    \caption{\textbf{Ablation study on the Parti-Prompts} dataset comparing the correlation to human preferences under varying factors: (a) the ViT training dataset, (b) input image size, (c) model capacity, (d) the number of features in the last ViT layer, and (e) zero-shot accuracy on ImageNet-1K.}
    \label{fig:parti_prompts_ablation}
    \vspace{-0.15in}
\end{figure*}

\noindent{\bf Effect of the size of the pre-training dataset on cFreD.}
\cref{fig:subfig_data_size} reports the \emph{partial-dependence} of \metric\ on dataset size after controlling all other covariates.  
Models trained on $\,{>}10$\,B images set the performance ceiling: the mixed-effects fit predicts $0.04$–$0.10$ higher Spearman $\rho$ than \emph{every} smaller bucket (Type-II ANOVA $F{=}206$, $p{<}10^{-170}$).  
All sub-billion regimes have sizable drops, with the 10M-image bucket suffering the largest penalty (–0.102 OLS; –0.077 mixed).  
Once these controlled effects are considered, the seemingly “strong’’ 10M result in the raw plot disappears and the relation between data scale and alignment becomes \emph{monotonic}.  
These findings confirm that sheer scale, rather than subtle confounds such as architecture or resolution, remains the single most powerful driver of \metric’s agreement with human judgments, underscoring the value of massive and well-curated corpora for text-conditioned image generation metrics.

\vspace{0.03in}
\noindent{\bf Effect of image size on cFreD.} 
\cref{fig:subfig_img_size} plots the \emph{partial-dependence} of \metric\ on input crop size (shortest side, in pixels) after controlling for data scale, parameter counts, and feature width.  The controlled effect is sharply negative: $\beta=-2.0\times10^{-4}$ Spearman~$\rho$ per pixel (OLS $t=-29$, $p<10^{-100}$; mixed-effects $z=-4.3$, $p<10^{-4}$), and Type-II ANOVA tags image size as a major factor ($F{=}860$, $p<10^{-154}$).  An increase in resolution from $224\!\rightarrow\!448$  is predicted to lower correlation by $\approx0.045$; this holds even larger inputs (e.g., $\geq\!896$), incurring steeper penalties.  Moderate resolutions remain competitive—Inception-scale crops land near the peak; and the DINOv2 $518$\,px setting lies on the declining slope.  Apparent non-monotonicities in the raw means largely vanish once covariates are held fixed, suggesting that oversized crops dilute semantically salient regions relative to their captions (higher resolution images would need more detailed captions), degrading alignment with human judgments.

\vspace{0.03in}
\noindent{\bf Effect of ViT model size on cFreD.}
\cref{fig:subfig_model_size} plots the \emph{partial-dependence} of \metric\ on image-encoder parameter count (log scale).  Controlling for data scale, crop size, and feature width reverses the naive trend: each $10{\times}$ increase in parameters \emph{decreases} Spearman~$\rho$ by $1.8$ points (OLS $\beta=-0.018$, $t=-8.0$, $p=2{\times}10^{-15}$) and by $2.6$ points in the mixed model ($z=-2.5$, $p=0.011$).  Type-II ANOVA corroborates the effect ($F{=}64$, $p<10^{-14}$).  Hence, once confounds are removed, oversized ViT backbones offer no benefit, and can even hurt, alignment with human preferences, suggesting that additional capacity is unnecessary for \metric.

\vspace{0.03in}
\noindent{\bf Effect of ViT Feature dimensionality on cFreD.}
\cref{fig:subfig_num_features} shows the \emph{partial-dependence} of \metric\ on ViT embedding dimensionality.  After controlling for data scale, crop size, and parameter count, each additional $+100$ channels boosts Spearman~$\rho$ by only $+0.0025$ (OLS $\beta=2.5{\times}10^{-5}$, $t=5.1$, $p=3.6{\times}10^{-7}$; mixed $z=17.4$, $p<10^{-15}$).  Gains taper beyond $\sim\!1\,\mathrm{k}$ dims but never reverse, contrary to the apparent high-dimensional downturn in the raw plot.  Widths in the $1,024$–$1,280$ range maximize correlation, yet leaner heads (512–640 dims) lag by barely $<\!0.01$.  Thus, modest widening is beneficial, but very large feature spaces deliver only marginal improvements relative to their computational cost.

\vspace{0.03in}
\noindent{\bf Effect of Zero-Shot ImageNet Accuracy on cFreD.}
~\cref{fig:subfig_zs_in1k} shows a boxplot of the correlation between \metric and human preferences as a function of zero-shot ImageNet accuracy, evaluated exclusively on image-text pretrained models~\cite{radford2021learning, schuhmann2022laion, fang2023data, gadre2023datacomp, xu2023demystifying}. Higher zero-shot accuracies generally correspond to stronger correlations with human judgments, though variance exists within each accuracy level. Models with the highest accuracy (81.37\%) achieve the highest median correlation, while models at the lower end (51.53\%) cluster around lower correlation values. These findings suggest that zero-shot ImageNet performance can serve as a useful indicator of downstream alignment with human preferences despite considerable variability among models with similar accuracy levels.

\subsection{Resilience to distortions}
Here, we provide additional evidence that FID detects failures of image-text alignment under \emph{complex} perturbations. For reference, it was shown in~\cite{heusel2017gans} that FID accurately captures image distortions under low-level image processing distortions such as Gaussian noise and Gaussian blur. Since \metric is a generalization of FID, it also inherits this resilience to such distortions. Therefore, we instead focus on two harder stress tests: (i) disrupting the \emph{image} in latent space and (ii) disrupting the \emph{text} conditioning signal.

\noindent\textbf{Image distortions.}\;
We follow the same setting as~\cite{szegedy2016rethinking} and first encode every generated image with the pretrained Cosmos tokenizer~\cite{agarwal2025cosmos}, randomly replace a fixed fraction of latent tokens with codes uniformly sampled from the same codebook, and then decode the corrupted latents back. This procedure preserves the global colour statistics of the training data while progressively scrambling high-level structure.  As shown in \cref{fig:subfig_img_distortions}, \metric\ \emph{monotonically increases} with the corruption ratio, faithfully mirroring the visual degradation.  In contrast, FID \emph{decreases} between 5--20\% corruption, suggesting that the images have \emph{improved}, before shooting up once the latent codes become overwhelmingly noisy.  The non-monotonic behaviour arises because replacing a moderate number of latents can accidentally move samples \emph{closer} to the marginal distribution of the real data, something FID cannot disambiguate without joint image–text statistics.  \metric, by cross-checking against captions, correctly penalises the emerging misalignment at every corruption level.

\noindent\textbf{Text distortions.}\;
We corrupt captions fed to the text encoder while keeping images fixed, using three ways:
(i) \textsc{MASK}: random token masking; 
(ii) \textsc{DROP}: random token deletion; 
and (iii) \textsc{HALLU}: hallucinating a portion of the caption using
(\texttt{LLaMA-3.1-70B-instruct}~\cite{dubey2024llama}).
The trends (\cref{fig:subfig_txt_distortions}) reveal three regimes.  
\emph{Low corruption:} \metric\ remains well above the constant FID baseline, signalling correct alignment.  
\emph{Medium corruption:} for \textsc{MASK} and \textsc{DROP}, the curves drift steadily downward and converge to FID, exactly as predicted by our formulation. When the text embeddings become noisy, the cross-covariance term averages out and \metric\ collapses to the image-only score.  
\emph{Hallucination:} hallucinated captions drive \metric\ \emph{below} FID, indicating anti-alignment between the fabricated text and the true image content.  Throughout, FID fails to detect, as it only sees unchanged pixels.

\section{Discussion and Conclusion}
\label{sec:conclusion}
 
In this work, we extend and repurpose Conditional Fr\'echet Inception Distance ($\mathrm{cFID}$), first introduced in~\cite{soloveitchik2021conditional}, for text-to-image and text-to-video evaluation by replacing outdated backbones (InceptionV3) with modern vision and text encoders. 
Our analysis shows that transformer-based encoders (e.g., ViT, DINOv2, CLIP) align significantly better with human preferences. 
Building on this, we introduce \metric, which captures semantic consistency between text and image, unlike FID or IS, and achieves higher correlation with human judgments. 
Our ablation further shows that performance depends on balancing model complexity, data diversity, and representation quality, not scaling the model or dataset.
We advocate \metric as a robust, plug-and-play metric requiring no annotations or training.
By open-sourcing our implementation, we aim to promote more systematic evaluations and accelerate generative model research.

\metric also offers a promising extension.
\metric can incorporate human preference-trained models' encoders as backbones for hybrid evaluation, which leverages \metric's robust semantic alignment and human preference data's fine-grained insights. It can also be adapted to other conditional tasks like image captioning or audio-to-video generation.
Furthermore, \metric can be generalized to evaluating multi-conditional models, such as ControlNet~\cite{zhang2023adding}, making it broadly applicable across generative tasks.

{
    \small
    \bibliographystyle{ieeenat_fullname}
    \bibliography{sections/6_references}
}

\appendix
\counterwithin{figure}{section}
\counterwithin{table}{section}
\renewcommand\thefigure{\thesection.\arabic{figure}}
\renewcommand\thetable{\thesection.\arabic{table}} 

\section{Image and Text Backbone Models}
In our experiments, we used 46 different image backbone models and 43 different text backbone models. For vision models, we employed those trained with self-supervised learning, including ViT trained on ImageNet-1k~\cite{dosovitskiy2020image}, ViT trained on ImageNet-21k~\cite{steiner2021train}, MAE~\cite{he2022masked}, DINOv2~\cite{oquab2023dinov2}, MoCOv3~\cite{chen2021empirical}, and I-JEPA trained on both ImageNet-1k and ImageNet-22k~\cite{assran2023self}. We also incorporated image-and-text-aligned pretrained models such as CLIP~\cite{radford2021learning}, MetaCLIP~\cite{xu2023demystifying}, DFN-CLIP~\cite{fang2023data}, OpenCLIP~\cite{cherti2023reproducible}, DataComp-CLIP~\cite{gadre2023datacomp}, ConvNeXT-CLIP~\cite{schuhmann2022laion}, EVA02~\cite{fang2023eva}, and SigLIP~\cite{zhai2023sigmoid}. Additional image models included SAM-ViT~\cite{kirillov2023segment} and Inception V3~\cite{szegedy2016rethinking}.
For text models, we employed autoencoding models such as  RoBERTa~\cite{liu2019roberta}, BERT~\cite{devlin2019bert}, ALBERT~\cite{lan2019albert}, ModernBERT~\cite{warner2024smarter}, XLM-RoBERTa~\cite{conneau2019unsupervised}, as well as sequence-to-sequence models including FLAN-T5~\cite{chung2024scaling} and T5~\cite{raffel2020exploring} were used. We also used the text encoders from the aforementioned image-and-text-aligned pretrained models.
The complete lists of image and text backbone models used in our experiments are presented in Tables \ref{tab:image_model_list} and \ref{tab:text_model_list}, respectively.

\section{Results on Text-to-Image}
We provide in-depth results on three text-to-image benchmarks: HPDv2 (\cref{tab:HPSv2_test_rank}), Parti-Prompts (\cref{tab:partiprompt_rank}), and a random selection of COCO prompts (\cref{tab:coco_main}). 
\begin{table*}[ht!]
\centering
\footnotesize
\setlength{\tabcolsep}{3pt}
\renewcommand{\arraystretch}{1.2}
%\resizebox{\textwidth}{!}{
\begin{tabular}{lrrrrrrrrrrrr c rrrrrrrr}
\toprule

\multirow{3}{*}{\textbf{Models}} & \multicolumn{12}{c}{\textbf{Statistical Metric}}                              && \multicolumn{8}{c}{\textbf{Human Preference Trained Metric}}

\\ \cmidrule(l){2-13} \cmidrule(l){15-22}

& \multicolumn{2}{c}{\textbf{Human}$\uparrow$} & 
\multicolumn{2}{c}{\textbf{FID}$\downarrow$} & 
\multicolumn{2}{c}{$\text{\textbf{FD}}_{\text{\tiny{\textbf{DINOv2}}}}\downarrow$} & 
\multicolumn{2}{c}{\textbf{CLIP}$\uparrow$} &  
\multicolumn{2}{c}{\textbf{CMMD}$\downarrow$} & 
\multicolumn{2}{c}{\textbf{cFreD}$\downarrow$} &&
\multicolumn{2}{c}{ \textbf{Aesthetic}$\uparrow$} &
\multicolumn{2}{c}{\textbf{ImReward}$\uparrow$} & 
\multicolumn{2}{c}{\textbf{HPS} v2$\uparrow$} & 
\multicolumn{2}{c}{\textbf{MPS}$\uparrow$} 

\\ \cmidrule(l){2-3} \cmidrule(l){4-5} \cmidrule(l){6-7} \cmidrule(l){8-9} \cmidrule(l){10-11} \cmidrule(l){12-13} \cmidrule(l){15-16} \cmidrule(l){17-18} \cmidrule(l){19-20} \cmidrule(l){21-22}

& \textbf{R\#}    & \textbf{Rate}     
& \textbf{R\#}   & \textbf{Score}    
& \textbf{R\#} & \textbf{Score}       
& \textbf{R\#}    & \textbf{Score}    
& \textbf{R\#}  & \textbf{Score}   
& \textbf{R\#}  & \textbf{Score}   
&& \textbf{R\#}  & \textbf{Score}   
& \textbf{R\#}   & \textbf{Score}   
& \textbf{R\#} & \textbf{Score}  
& \textbf{R\#} & \textbf{Score}  \\ 

\midrule
\textit{GLIDE}~\cite{nichol2021glide}       & 1       & 80.87\%   & 1     & 7.90    & 1      & 6.88     & 9      & 14.34     & 1     & 2.42    & 1     & 3.79    && 1     &  5.55   & 3      & 0.37  & 2    & 25.52  & 1    & 12.72  \\

\textit{COCO}~\cite{lin2014microsoft}        & 2       & 80.66\%   & 7     & 13.11   & 7      & 13.05    & 10     & 13.11     & 5     & 15.07   & 4     & 4.55    && 5     &  5.03   & 1      & 0.55  & 1    & 25.64  & 5    & 10.92  \\

\textit{FuseDream}~\cite{liu2021fusedream}  & 3       & 76.29\%   & 2     & 8.39    & 2      & 7.59     & 5      & 15.07     & 4     & 5.41    & 2     & 4.16    && 4     &  5.34   & 2      & 0.47  & 3    & 24.40  & 2    & 12.44 \\

\textit{DALLE~2}~\cite{ramesh2022hierarchical}    & 4       & 75.87\%   & 3     & 9.16    & 3      & 7.95     & 8      & 14.39     & 3     & 4.06    & 3     & 4.42    && 3     &  5.40   & 6      & 0.07  & 5    & 23.81  & 4    & 11.75    \\

\textit{VQGAN+CLIP}~\cite{esser2021taming}  & 5       & 68.78\%   & 4     & 10.11   & 4      & 8.70     & 7      & 14.41     & 2     & 3.70    & 5     & 4.90    && 2     &  5.40   & 5      & 0.08  & 4    & 23.93  & 3    & 11.82  \\ 

\textit{CogView2}~\cite{ding2022cogview2}    & 6       & 39.00\%   & 6     & 12.65   & 6      & 12.87    & 3      & 15.45     & 8     & 45.64   & 7     & 6.93    && 7     &  4.82   & 7      & 0.02  & 7    & 19.45  & 7    & 8.85    \\

\textit{SDv1.4}~\cite{rombach2022high}    & 7       & 38.36\%   & 5     & 12.51   & 5      & 11.93    & 4      & 15.42     & 6     & 28.52   & 8     & 7.18    && 8     &  4.56   & 8      & -0.67 & 8    & 19.44  & 8    & 8.00    \\

\textit{VQ-Diffusion}~\cite{gu2022vector} & 8      & 32.04\%   & 8     & 13.85   & 8      & 13.12    & 6      & 14.71     & 7     & 33.40   & 6     & 6.59    && 6     &  4.88   & 4      & 0.17  & 6    & 21.91  & 6    & 10.15  \\ 

\textit{SDv2.0}~\cite{rombach2022high}     & 9       & 22.00\%   & 9     & 14.74   & 9      & 14.23    & 2      & 15.62     & 10    & 55.88   & 9     & 8.16    && 9     &  4.54   & 9      & -0.72 & 9    & 18.45  & 9    & 7.24   \\ 

\textit{LAFITE}~\cite{zhou2022towards}     & 10      & 9.07\%    & 10    & 15.12   & 10     & 14.63    & 1      & 16.01     & 9     & 53.22   & 10    & 9.06    && 10    &  4.23   & 10     & -1.45 & 10   & 15.03  & 10   & 5.08  \\ 
 
\midrule
\multicolumn{1}{c}{$\bm{\rho}^2$}  & 
\multicolumn{2}{c}{-}  &  
\multicolumn{2}{c}{0.70} & 
\multicolumn{2}{c}{0.65} & 
\multicolumn{2}{c}{0.63} &  
\multicolumn{2}{c}{0.88} & 
\multicolumn{2}{c}{\textbf{0.97}} && 
\multicolumn{2}{c}{0.83}                & 
\multicolumn{2}{c}{0.71}  & 
\multicolumn{2}{c}{\underline{0.90}} & 
\multicolumn{2}{c}{0.86}         \\ 
%\midrule
\multicolumn{1}{c}{\textbf{Rank Acc.}} & 
\multicolumn{2}{c}{-} & 
\multicolumn{2}{c}{86.7} & 
\multicolumn{2}{c}{86.7} & 
\multicolumn{2}{c}{15.6}   &  
\multicolumn{2}{c}{80.0} & 
\multicolumn{2}{c}{\textbf{91.1}} && 
\multicolumn{2}{c}{82.2}                &
\multicolumn{2}{c}{84.4}   & 
\multicolumn{2}{c}{\underline{88.9}}    & 
\multicolumn{2}{c}{86.7}        \\\bottomrule
\end{tabular}
%}
\caption{Text-to-image model ranking and scores by statistical models (FID, $\text{FD}_{\text{\tiny{DINOv2}}}$, CLIP score, CMMD, and cFreD) and models that were trained with human preference (Aesthetic Score, ImageReward, HPSv2, and MPS) on HPDv2 test set. Best results in \textbf{bold}, second best \underline{underlined}}
\label{tab:HPSv2_test_rank}
\end{table*}

\begin{table*}[ht!]
\centering
\footnotesize
\setlength{\tabcolsep}{3pt}
\renewcommand{\arraystretch}{1.2}
%\resizebox{\textwidth}{!}{
\begin{tabular}{lcccccccccccc c cccccccc}
\toprule

\multirow{3}{*}{\textbf{Models}} & 

\multicolumn{12}{c}{\textbf{Statistical Metric}}                         && \multicolumn{8}{c}{\textbf{Human Preference Trained Metric}}   

\\ \cmidrule(l){2-13} \cmidrule(l){15-22} 

& \multicolumn{2}{c}{\textbf{Human}$\uparrow$} & 
\multicolumn{2}{c}{\textbf{FID}$\downarrow$} & 
\multicolumn{2}{c}{$\text{\textbf{FD}}_{\text{\tiny{\textbf{DINOv2}}}}\downarrow$} & 
\multicolumn{2}{c}{\textbf{CLIP}$\uparrow$} &  
\multicolumn{2}{c}{\textbf{CMMD}$\downarrow$} & 
\multicolumn{2}{c}{\textbf{cFreD}$\downarrow$} &&
\multicolumn{2}{c}{ \textbf{Aesthetic}$\uparrow$} &
\multicolumn{2}{c}{\textbf{ImReward}$\uparrow$} & 
\multicolumn{2}{c}{\textbf{HPS} v2$\uparrow$} & 
\multicolumn{2}{c}{\textbf{MPS}$\uparrow$} 

\\ \cmidrule(l){2-3} \cmidrule(l){4-5} \cmidrule(l){6-7} \cmidrule(l){8-9} \cmidrule(l){10-11} \cmidrule(l){12-13} \cmidrule(l){15-16} \cmidrule(l){17-18} \cmidrule(l){19-20} \cmidrule(l){21-22}

& \textbf{R\#}    & \textbf{Rate}     
& \textbf{R\#}   & \textbf{Score}    
& \textbf{R\#} & \textbf{Score}      
& \textbf{R\#}    & \textbf{Score}    
& \textbf{R\#}  & \textbf{Score}   
& \textbf{R\#}  & \textbf{Score}  
&& \textbf{R\#}  & \textbf{Score}   
& \textbf{R\#}   & \textbf{Score}   
& \textbf{R\#} & \textbf{Score}  
& \textbf{R\#} & \textbf{Score}  

\\ \midrule
\textit{SDXL}~\cite{podell2023sdxl}       & 1       & 69.84\%   & 1     & 31.24    & 1      & 23.50     & 2      & 32.79     & 1     & 2.55    & 1     & 2.98    && 3     &  5.64   & 1      & 0.95  & 1    & 28.63  & 4    & 10.33     \\

\textit{Kand2}~\cite{kandinsky}       & 
2       & 46.10\%   & 2     & 32.08    & 2      & 24.35     & 3      & 32.62     & 2     & 2.77    & 2     & 3.21    && 2     &  5.65   & 2      & 0.90  & 2    & 28.13  & 2    & 11.29   \\

\textit{Wuerst}~\cite{pernias2024wrstchen}      & 3       & 42.68\%   & 3     & 38.43    & 3      & 30.93     & 4      & 31.72     & 3     & 3.83    & 3     & 4.18    && 1     &  5.71   & 3      & 0.79  & 3    & 27.79  & 1    & 11.30  \\
\textit{Karlo}~\cite{kakaobrain2022karlo}       & 4       & 29.21\%   & 4     & 48.40    & 4      & 41.57     & 1      & 33.01     & 4     & 19.95   & 4     & 5.45    && 4     &  4.93   & 4      & 0.70  & 4    & 26.56  & 3    & 11.11         
 \\\midrule
 
\multicolumn{1}{c}{$\bm{\rho}^2$} & 
\multicolumn{2}{c}{-}  &  
\multicolumn{2}{c}{0.70} & 
\multicolumn{2}{c}{0.70} & 
\multicolumn{2}{c}{0.12} &  
\multicolumn{2}{c}{0.54} & 
\multicolumn{2}{c}{0.73} && 
\multicolumn{2}{c}{0.43} & 
\multicolumn{2}{c}{\underline{0.81}}  & 
\multicolumn{2}{c}{\textbf{0.83}} & 
\multicolumn{2}{c}{0.65}         \\ %\midrule
%Spearman $\rho$ to Human Eval.  & \multicolumn{2}{c}{-} & \multicolumn{2}{c}{1.00} & \multicolumn{2}{c}{1.00} & \multicolumn{2}{c}{0.04}   &  \multicolumn{2}{c}{1.00} & \multicolumn{2}{c}{1.00}\vline & \multicolumn{2}{c}{0.04} & \multicolumn{2}{c}{1.00}   & \multicolumn{2}{c}{1.00}    & \multicolumn{2}{c}{0.}        \\ 
\bottomrule
\end{tabular}
%}
\caption{Text-to-image model ranking and scores by statistical metrics (FID, $\text{FD}_{\text{\tiny{DINOv2}}}$, CLIP score, CMMD, and \metric) and models that were trained with human preference (Aesthetic Score, ImageReward, and MPS) on Parti-Prompt. Best results in \textbf{bold}, second best \underline{underlined}}
\label{tab:partiprompt_rank}
\end{table*}
\begin{table*}[ht!]
\centering
\footnotesize
\setlength{\tabcolsep}{3pt}
\renewcommand{\arraystretch}{1.2}
%\resizebox{\textwidth}{!}{
\begin{tabular}{lrrrrrrrrrrrr c rrrrrrrr}
\toprule

\multirow{3}{*}{\textbf{Models}} & 
\multicolumn{12}{c}{\textbf{Statistical Metric}} &&
\multicolumn{8}{c}{\textbf{Human Preference Trained Metric}}                     
\\ 

\cmidrule(l){2-13} \cmidrule(l){15-22} 

%& \multicolumn{2}{c}{\textbf{\scriptsize Human Eval.}$\uparrow$} & 
& \multicolumn{2}{c}{\textbf{\scriptsize Humans}$\uparrow$} & 
\multicolumn{2}{c}{\textbf{\scriptsize FID}$\downarrow$} & 
\multicolumn{2}{c}{$\text{\scriptsize \textbf{FD}}_{\text{\tiny{\textbf{DINOv2}}}}\downarrow$} & 
\multicolumn{2}{c}{\textbf{\scriptsize CLIP}$\uparrow$} &  
\multicolumn{2}{c}{\textbf{\scriptsize CMMD}$\downarrow$} & 
\multicolumn{2}{c}{\textbf{cFreD}$\downarrow$} &&
\multicolumn{2}{c}{ \textbf{\scriptsize Aesthetic}$\uparrow$} &
\multicolumn{2}{c}{\textbf{\scriptsize ImReward}$\uparrow$} & 
\multicolumn{2}{c}{\textbf{\scriptsize HPS v2}$\uparrow$} & 
\multicolumn{2}{c}{\textbf{\scriptsize MPS}$\uparrow$} 

\\ 
\cmidrule(l){2-3} 
\cmidrule(l){4-5} 
\cmidrule(l){6-7} 
\cmidrule(l){8-9} 
\cmidrule(l){10-11} 
\cmidrule(l){12-13} 
\cmidrule(l){15-16} 
\cmidrule(l){17-18} 
\cmidrule(l){19-20} 
\cmidrule(l){21-22}
& 
\textbf{\scriptsize R\#}    &  \textbf{\scriptsize ELO}     & 
\textbf{\scriptsize R\#}   & \textbf{\scriptsize Score}    & 
\textbf{\scriptsize R\#} & \textbf{\scriptsize Score}       & 
\textbf{\scriptsize R\#}    & \textbf{\scriptsize Score}    & 
\textbf{\scriptsize R\#}  & \textbf{\scriptsize Score}   & 
\textbf{\scriptsize R\#}  & \textbf{\scriptsize Score}   && 
\textbf{\scriptsize R\#}  & \textbf{\scriptsize Score}   & 
\textbf{\scriptsize R\#}   & \textbf{\scriptsize Score}   & 
\textbf{\scriptsize R\#} & \textbf{\scriptsize Score}  & 
\textbf{\scriptsize R\#} & \textbf{\scriptsize Score}  \\ 

\midrule
\textit{FLUX.1[dev]}~\cite{flux2024}           
& 1       & 1083      & 5     & 10.45   & 7    & 7.21     & 9      & 30.72    & 8     & 6.08   &   4   &      9.93    && 
2     &  5.75   & 3      & 1.10    & 2    & 30.74  & 2    & 12.90    \\

\textit{SDv3.5-L Turbo}~\cite{esser2024scaling}  & 2       & 1073      & 9     & 11.68   & 9    & 8.12     & 6      & 31.11    & 7     & 48.52   & 7     & 10.44    && 
6     &  5.50   & 6      & 0.64    & 7    & 26.52  & 6    & 10.88  \\

\textit{SDv3.5-L}~\cite{esser2024scaling}       & 3       & 1069      & 2     & 10.27   & 4    & 6.77     & 1      & 31.74    & 4     & 40.27   & 2     & 9.49    && 
4     &  5.55   & 2      & 1.10    & 3    & 30.07  & 3    & 12.30    \\

\textit{Playgroundv2.5}~\cite{li2024playground}      & 4       & 997       & 7     & 10.90   & 8    & 7.50     & 7      & 31.03    & 9     & 66.10   & 5     & 10.08    && 
1     &  6.16   & 1      & 1.15    & 1    & 31.56  & 1    & 13.15   \\ 

\textit{SDv3-M}~\cite{esser2024scaling}         & 5       & 944       & 6     & 10.46   & 5    & 7.09     & 3      & 31.68    & 2     & 34.34   & 1    & 9.49    && 
7     &  5.45   & 4      & 1.08    & 4    & 29.80  & 9    & 2.35  \\ 

\textit{SDXL}~\cite{podell2023sdxl}            & 6       & 890       & 3     & 10.31   & 2    & 6.66     & 2      & 31.70    & 6     & 45.07   & 3     & 9.73    &&
3     &  5.61   & 5      & 0.76    & 5    & 28.34  & 4    & 12.08  \\ 

\textit{SDv2.1}~\cite{rombach2022high}              & 7       & 752       & 1     & 10.14   & 1    & 6.55     & 4      & 31.42    & 3     & 35.17   & 6     & 10.24    && 
5     &  5.52   & 8      & 0.41    & 6    & 26.58  & 5    & 10.90  \\ 

\textit{Janus Pro}~\cite{chen2025janus}            & 8       & 740       & 8     & 10.97   & 6    & 7.17     & 8      & 30.99    & 5     & 43.51   & 9     & 10.76    && 
9     &  5.33   & 7      & 0.57    & 8    & 26.22  & 7    & 10.70 \\

\textit{SDv1.5}~\cite{rombach2022high}              & 9       & 664       & 4     & 10.41   & 3    & 6.73     & 5      & 31.21    & 1     & 29.89   & 8     & 10.58   &&
8     &  5.34   & 9      & 0.19    & 9    & 26.15  & 8    & 10.50   
 \\\midrule
\multicolumn{1}{c}{$\bm{\rho}^2$}  & 
\multicolumn{2}{c}{-}  &  
\multicolumn{2}{c}{0.08} & 
\multicolumn{2}{c}{0.29} & 
\multicolumn{2}{c}{0.00} &  
\multicolumn{2}{c}{0.22} & 
\multicolumn{2}{c}{0.33} &&
\multicolumn{2}{c}{0.27} &
\multicolumn{2}{c}{\textbf{0.69}}  & 
\multicolumn{2}{c}{\underline{0.48}} & 
\multicolumn{2}{c}{\underline{0.48}}         \\ %\midrule
\multicolumn{1}{c}{\textbf{Rank Acc.}}   & 
\multicolumn{2}{c}{-} & 
\multicolumn{2}{c}{41.67} & 
\multicolumn{2}{c}{36.11} & 
\multicolumn{2}{c}{47.22}   &  
\multicolumn{2}{c}{30.56} & 
\multicolumn{2}{c}{66.67} && 
\multicolumn{2}{c}{\underline{72.22}} & 
\multicolumn{2}{c}{\textbf{80.56}}   & 
\multicolumn{2}{c}{\textbf{80.56}}    & 
\multicolumn{2}{c}{\textbf{80.56}}        
\\ \bottomrule
\end{tabular}
%}
\caption{Text-to-image model ranking by automatic models(FID, $\text{FD}_{\text{\tiny{DINOv2}}}$, CLIP score, CMMD, and cFreD) and models that were trained with human preference (Aesthetic Score, ImageReward, HPSv2, and MPS) on randomly sampled COCO prompts. Rank Acc. below 0.5 indicates there are more discordant pairs than concordant ones. Best results in \textbf{bold}, second best \underline{underlined}}
\label{tab:coco_main}
\end{table*}

\vspace{0.02in}
\noindent \textbf{Results on HPDv2. } Table~\ref{tab:HPSv2_test_rank} summarizes the rankings and scores for images generated by various models—including one real image—across multiple evaluation metrics. Notably, \metric achieves the highest alignment with human preferences, reaching a correlation of 0.97. Among statistical metrics, \metric attains the highest correlation and is comparable to HPSv2 (0.94), a model explicitly trained on human preferences. Given that HPSv2 was trained on the HPSv2 training set, which includes four models from the test set, and employed the same annotators, it inherently encodes specific human preference biases of the same setting. In contrast, \metric achieves comparable or superior correlation with human evaluation without any human preference training. These results demonstrate that \metric provides more reliable rankings across diverse models compared to standard automatic metrics and metrics trained explicitly on human preference data.

\noindent Table~\ref{tab:HPSv2_test_rank} also reports the rank accuracy scores on the HPDv2 test set. Among all evaluated metrics, \metric achieves the highest rank accuracy (91.1\%), highlighting its strong correspondence with human judgments. HPSv2 follows as the second-best metric with an accuracy of 88.9\%, while both FID and $\text{FD}_{\text{DINOv2}}$ obtain competitive scores of 86.7\%. Overall, although models trained with human preference data tend to align well with human judgments, \metric emerges as the most robust and reliable metric.

\vspace{0.03in}
\noindent \textbf{Result on PartiPrompts Arena. } Table~\ref{tab:partiprompt_rank} presents the rankings and scores of text-to-image models evaluated on the Parti-Prompt Arena using both statistical metrics and human preference-trained models. Among the statistical metrics, \metric achieves the highest correlation with human evaluations (0.73), with FID and $\text{FD}_{\text{\tiny{DINOv2}}}$ both reaching a correlation of 0.70. In contrast, the CLIP score shows a very low correlation (0.12) with human judgments. In the human preference trained category, HPSv2 has the strongest alignment, achieving the highest correlation (0.83), followed by ImageReward (0.81) and MPS (0.65). These results highlight that while \metric is a robust automatic metric, HPSv2 stands out as the most effective in capturing human evaluation trends in the PartiPrompts Arena.

\vspace{0.03in}
\noindent \textbf{Results on COCO. }\quad
Table~\ref{tab:coco_main} presents an evaluation on the COCO dataset using nine modern text-to-image models, with human preference rankings sourced from the Text-to-Image Leaderboard and expressed as ELO scores. Among statistical metrics (FID, $\text{FD}_{\text{\footnotesize{DINOv2}}}$, CLIP, CMMD, and our proposed \metric), only \metric exhibits a strong correlation with human preferences, achieving a correlation of 0.33 and a non-trivial rank accuracy of 66.67\%. This result places \metric as the third most aligned metric overall, surpassed only by the human preference–trained metrics ImageReward, HPSv2, and MPS. Notably, all other statistical metrics show considerably weaker alignment with ELO rankings and, as a result, inverted the rankings, resulting in a Rank Acc. below 0.5. These findings highlight that \metric is sensitive to both visual fidelity and prompt consistency, reinforcing its value as a practical, training-free alternative for benchmarking text-to-image generation.

\section{Analysis on Selecting Image and Text Backbone Models}
We provide an in-depth analysis of how different image and text models affect \metric. In our experiments, we tested all possible combinations of 43 different image and text models, focusing exclusively on ViT-based architectures and excluding SigLIP models with high-resolution options.

\subsection{Spearman Correlation}
We present Spearman correlation heatmaps that compare various text and image models across three distinct datasets: Parti-Prompts, HPDv2, and a random selection of COCO prompts.

\begin{figure*}[ht!]
    \centering
    \includegraphics[width=0.90\textwidth]{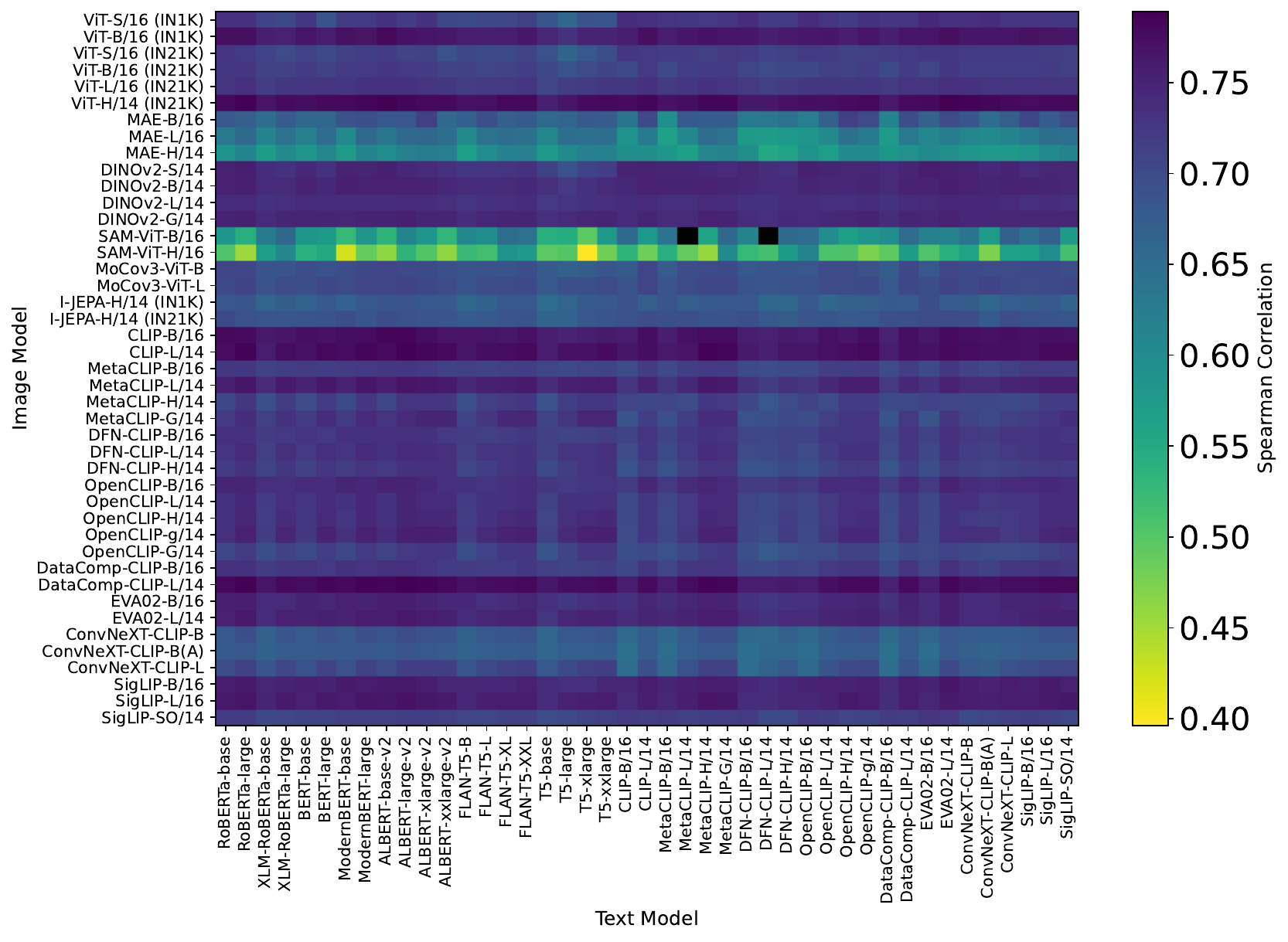}
    \caption{Spearman Correlation Heatmap on Parti-Prompts.}
    \label{fig:Parti_prompt_heatmap}
    \vspace{-0.05in}
\end{figure*}

~\cref{fig:Parti_prompt_heatmap} showcases the heatmap on Parti-Prompts. Notably, certain image models such as ViT-B/16 trained on ImageNet-1K and ViT-H/14 trained on ImageNet-21K show consistently high performance with different text models. In contrast, SAM-ViT-H/16 presents more variability with different text models. While most ViT, DINO, and CLIP-based models demonstrate strong correlations across different text models, MAE models show slightly lower correlations.

\begin{figure*}[t!]
    \centering
    \includegraphics[width=0.90\textwidth]{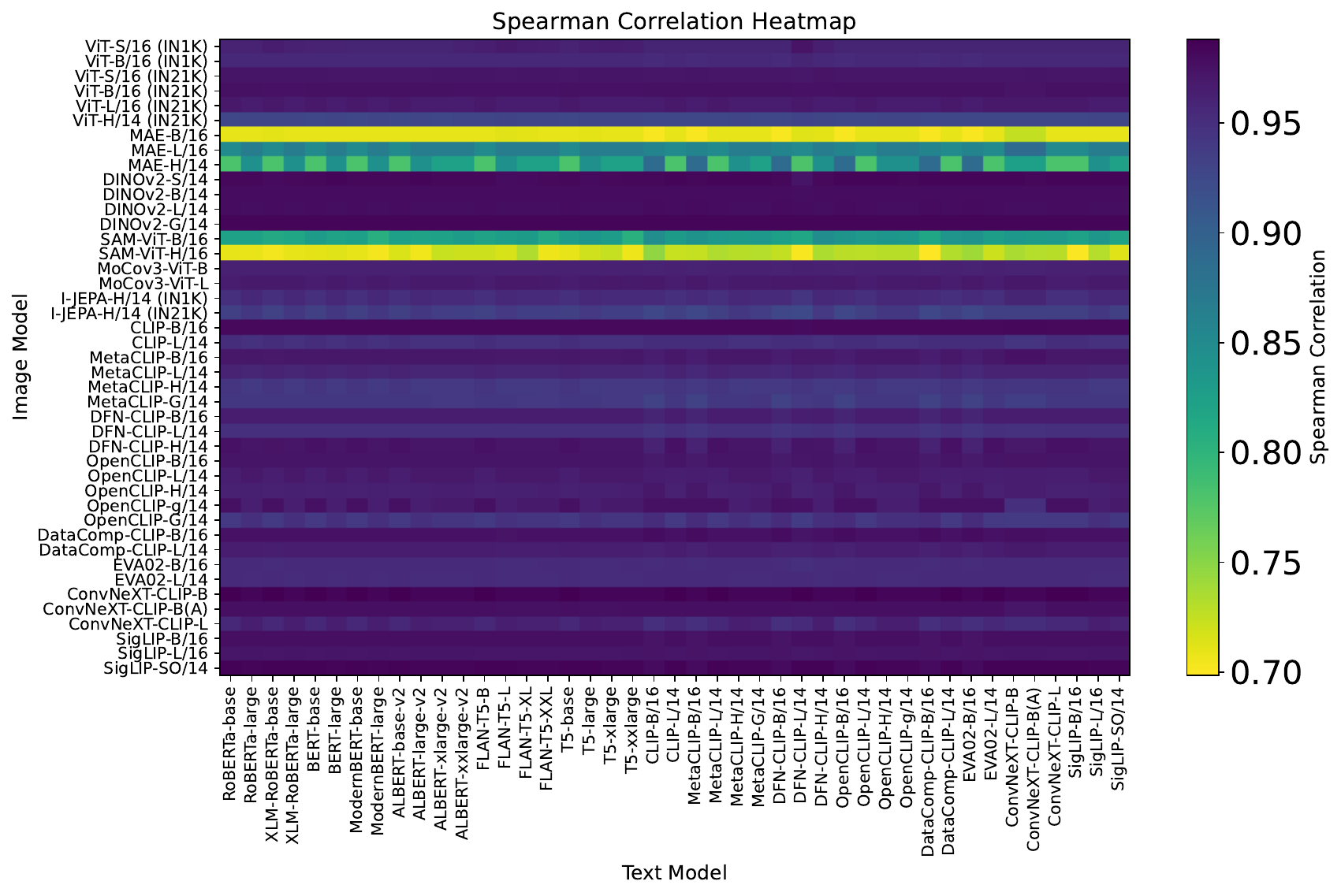}
    \caption{Spearman Correlation Heatmap on HPDv2.}
    \label{fig:HPDv2_heatmap}
    \vspace{-0.05in}
\end{figure*}

~\cref{fig:HPDv2_heatmap} presents the heatmap on HPDv2. Most models show strong correlations, whereas MAE models and SAM have lower correlations. Regardless of their performance levels, all image models show consistent correlation patterns across different text models.

\begin{figure*}[ht!]
    \centering
    \includegraphics[width=0.90\textwidth]{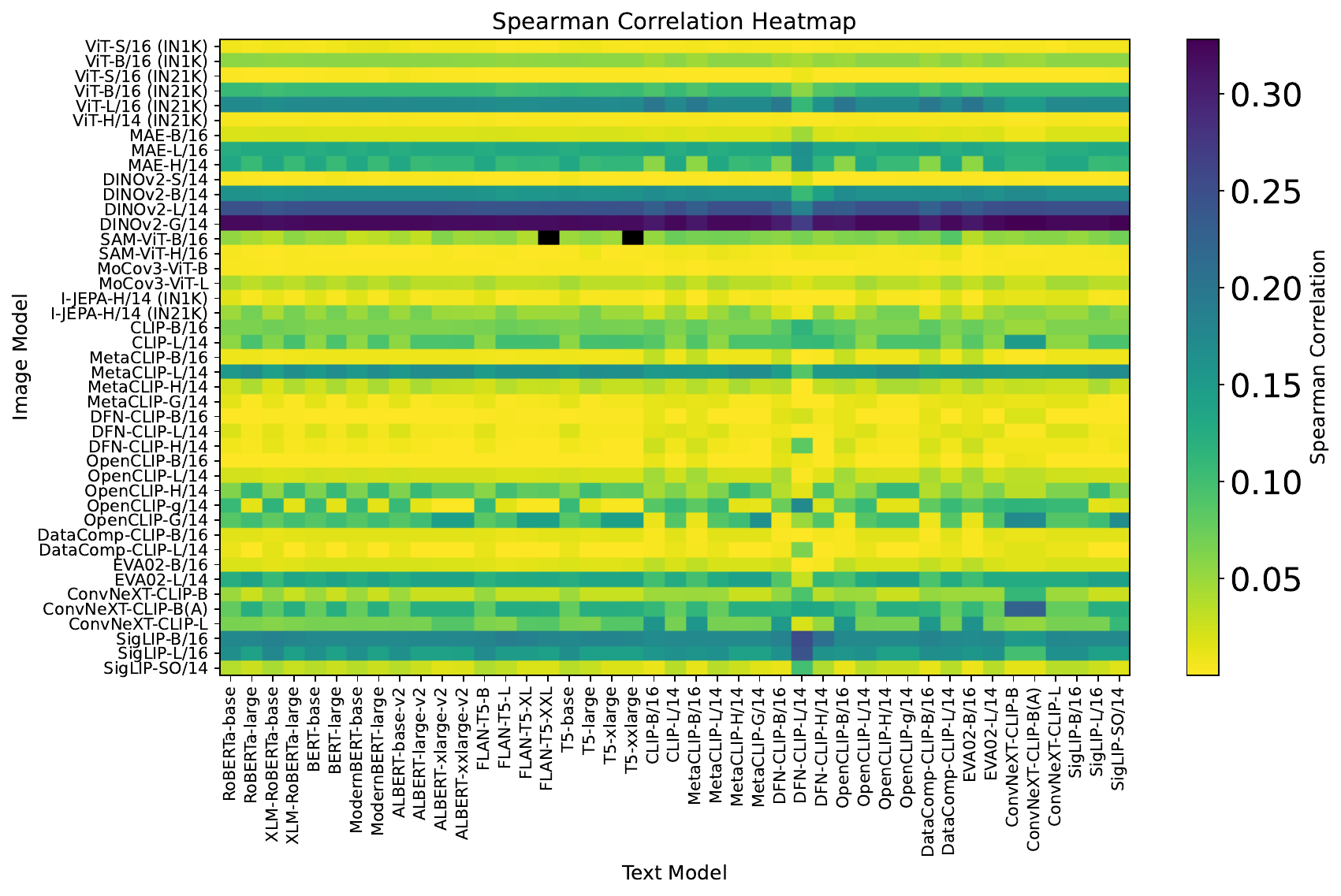}
    \caption{Spearman Correlation Heatmap on random COCO prompts.}
    \label{fig:COCO_heatmap}
    \vspace{-0.03in}
\end{figure*}

~\cref{fig:COCO_heatmap} provides the heatmap on randomly selected COCO prompts. The correlation values are significantly lower overall, ranging from approximately 0.00 to 0.30, which is much lower than both the Parti-Prompts and HPDv2 datasets. Most models show lower correlations, while only DINOv2 models consistently demonstrate stronger correlations. Compared to other datasets, this heatmap exhibits more variability, such as OpenCLIP showing a mixture of relatively higher and lower correlations depending on the text models.

These results highlight the importance of selecting compatible text and image models for improved cross-modal understanding, as not all combinations yield equally robust alignment. Additionally, it suggests that selecting a suitable image encoder plays a more pivotal role than choosing a text encoder, indicating that image encoder choice exerts a greater influence on overall performance.

\subsection{Effect of Visual Encoders on HPDv2}
In this section, we provide an analysis of how different visual encoder characteristics impact \metric's correlation with human preferences on HPDv2. For each factor examined, we report the average correlation across all possible text encoder combinations.

\noindent{\bf Effect of the size of the pre-training dataset on \metric.}
 ~\cref{fig:HPDv2_subfig_data_size} shows the correlation between \metric and human preferences as a function of the size of the pre-training dataset for Vision Transformer (ViT). In all ranges, it show high alignment with human judgments, with a correlation higher than 0.95 correlation across all data sizes. This indicates that factors beyond raw data quantity, such as diversity and quality, significantly influence performance.

\noindent{\bf Effect of image size on \metric.}
 ~\cref{fig:HPDv2_subfig_img_size} illustrates the correlation between \metric and human preferences across varying input image resolutions. We observe a nonmonotonic relationship: increasing resolutions do not consistently yield higher correlations. In particular, an image size from 224x224 to 896x896 all achieves a high correlation abovethan 0.96. 
 However, beyond 518x518, performance declines, reaching 0.964 at 896x896 and showing the lowest correlation of 0.774 at 1024x1024. 

\noindent{\bf Effect of ViT model size on \metric.}
The correlation between \metric and human preferences across Vision Transformer (ViT) sizes are presented in~\cref{fig:HPDv2_subfig_model_size}. 
It shows that all models achieve consistently high correlations—ranging from 0.945 to 0.986—indicating strong agreement with the target metric across scales. Interestingly, the \emph{SO} model attains the highest correlation at 0.986, while \emph{Gigantic} has the lowest, though still robust correlation of 0.945. The remaining models also cluster around correlations between 0.96 and 0.98, suggesting that simply increasing model size does not guarantee a strictly monotonic improvement in correlation.

\noindent{\bf Effect of ViT Feature dimensionality on \metric.}
 ~\cref{fig:HPDv2_subfig_num_features} shows the correlation between \metric and human preferences across ViT feature dimensions from 256 to 1664. The lowest number of the feature (256) shows the lowest correlation. However, we observe a clear plateau effect in performance once the feature count reaches 384, with correlation values stabilizing around 0.98-0.99 across a wide range of dimensionalities (512-1408). Interestingly, at extremely high dimensionalities (above 1536), we note a slight performance decline, with correlation dropping to 0.95 at 1664 features. This suggests an optimal range for feature dimensionality exists, beyond which additional computational complexity yields diminishing or even negative returns.

\noindent{\bf Effect of Zero-Shot ImageNet Accuracy on \metric.}
 ~\cref{fig:HPDv2_subfig_zs_in1k} depicts a boxplot of the correlation between \metric and human preferences as a function of zero-shot ImageNet accuracy, evaluated exclusively on image-text pretrained models~\cite{radford2021learning}. Higher zero-shot accuracies generally correspond to stronger correlations with human judgments, though variance exists within each accuracy bin. Interestingly, we find that correlations peak at a model with 66.58\% zero-shot accuracy and decrease as model accuracy gets higher.

\subsection{Effect of Visual Encoders on COCO prompts}
In this section, we provide an analysis of how different visual encoder characteristics impact \metric's correlation with human preferences on randomly selected COCO prompts. For each factor examined, we report the average correlation across all possible text encoder combinations.

\begin{figure*}[th]
    \centering
    % First row: three images
    \begin{subfigure}[b]{0.32\textwidth}
        \centering
        \includegraphics[width=\textwidth]{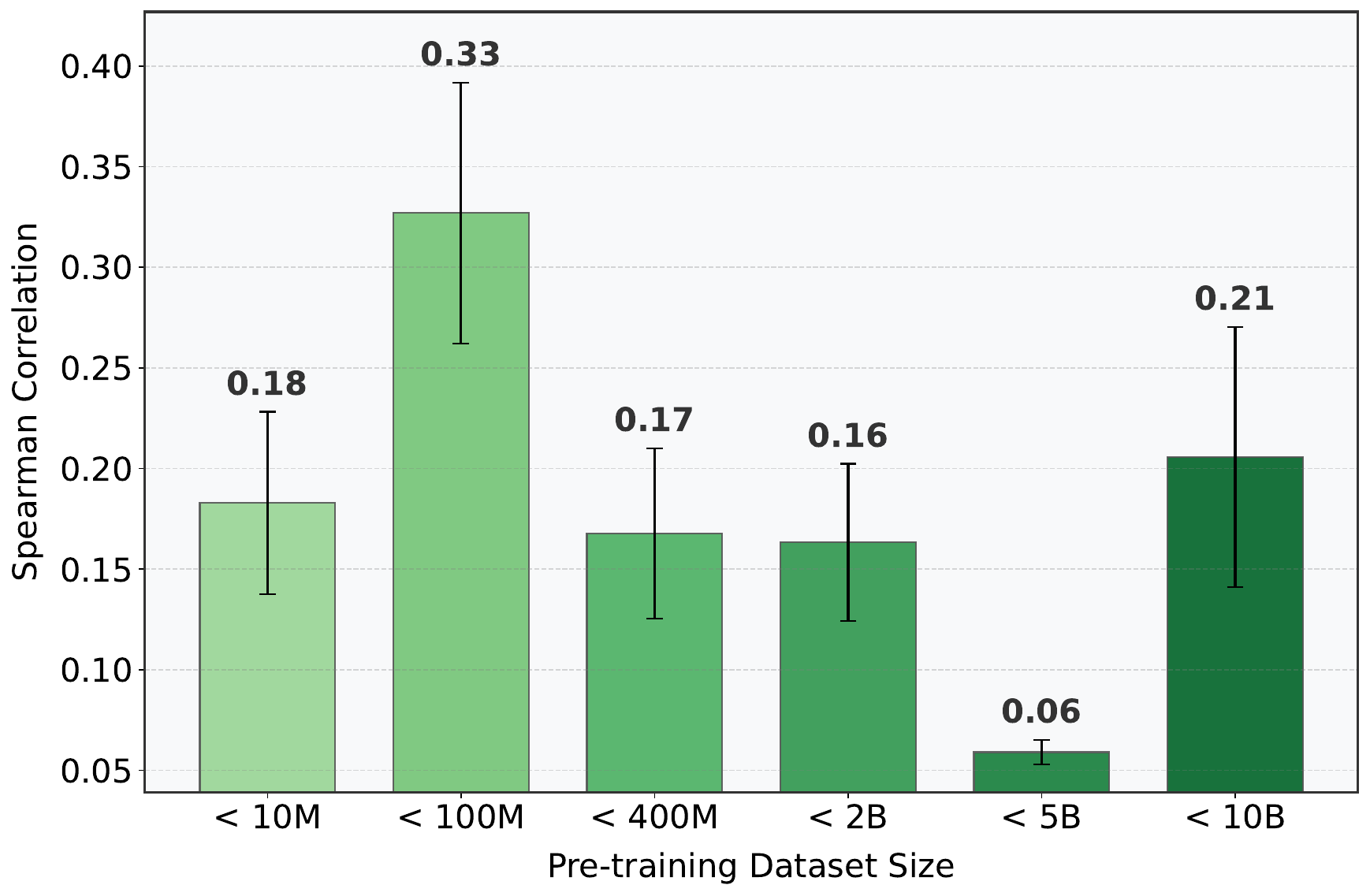}
        \caption{Effect of training data.}
        \label{fig:COCO_subfig_data_size}
    \end{subfigure}
    \hfill
    \begin{subfigure}[b]{0.32\textwidth}
        \centering
        \includegraphics[width=\textwidth]{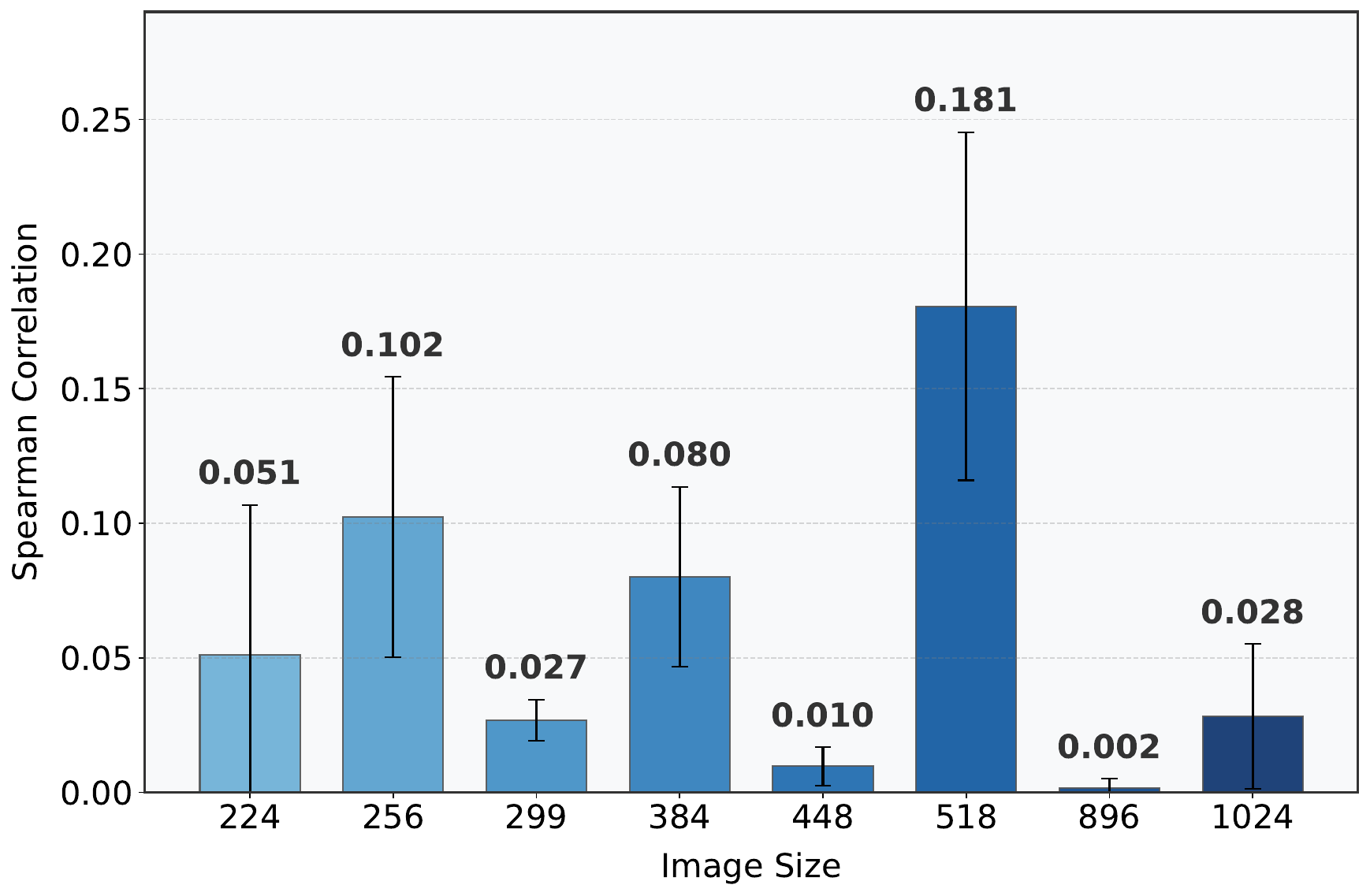}
        \caption{Effect of image size.}
        \label{fig:COCO_subfig_img_size}
    \end{subfigure}
    \hfill
    \begin{subfigure}[b]{0.32\textwidth}
        \centering
        \includegraphics[width=\textwidth]{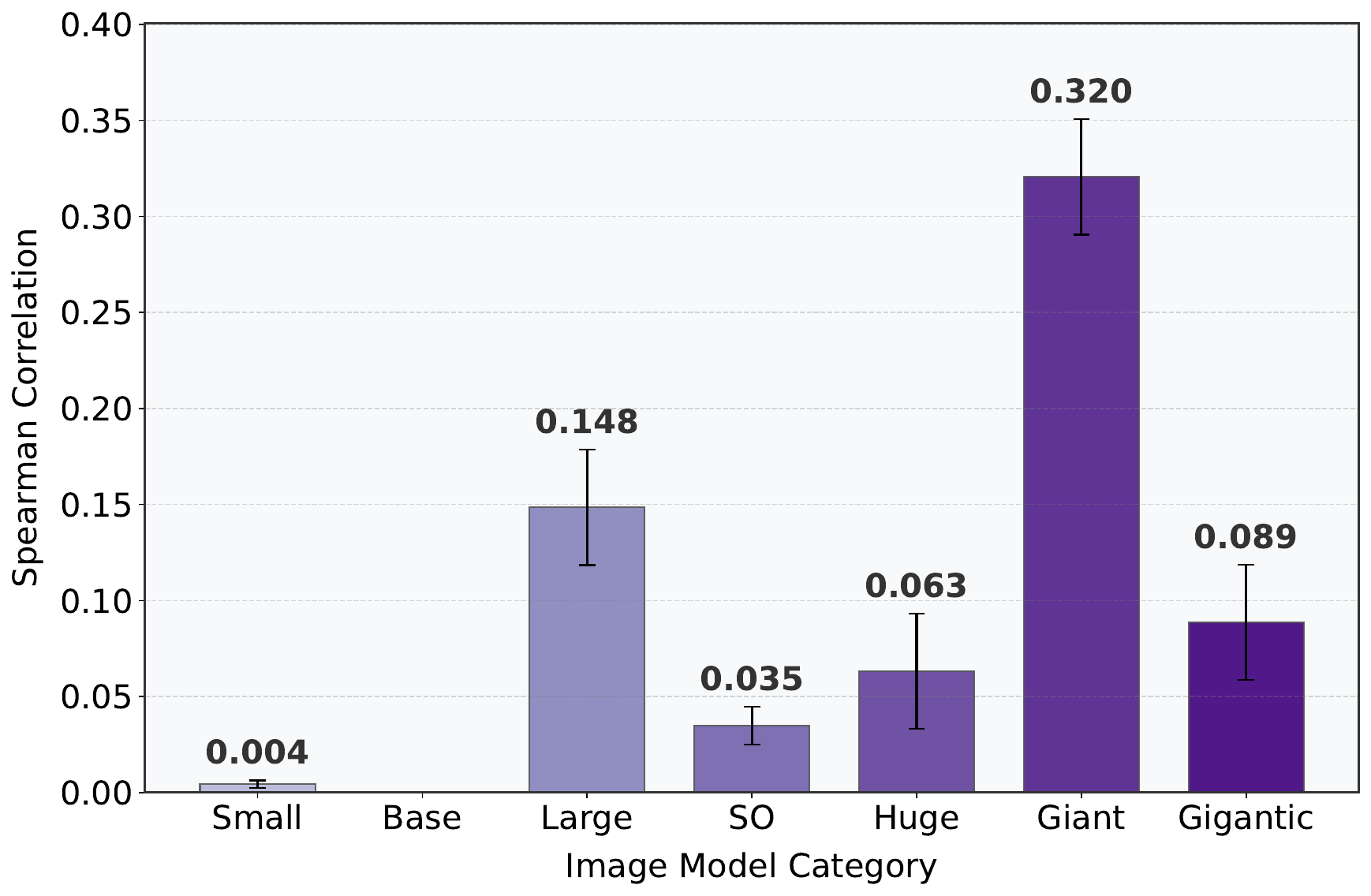}
        \caption{Effect of model size.}
        \label{fig:COCO_subfig_model_size}
    \end{subfigure}
    
    \vspace{1em}
    
    % Second row: two images centered under the three above
    \begin{minipage}{0.7\textwidth}
        \centering
        \begin{subfigure}[b]{0.46\textwidth}
            \centering
            \includegraphics[width=\textwidth]{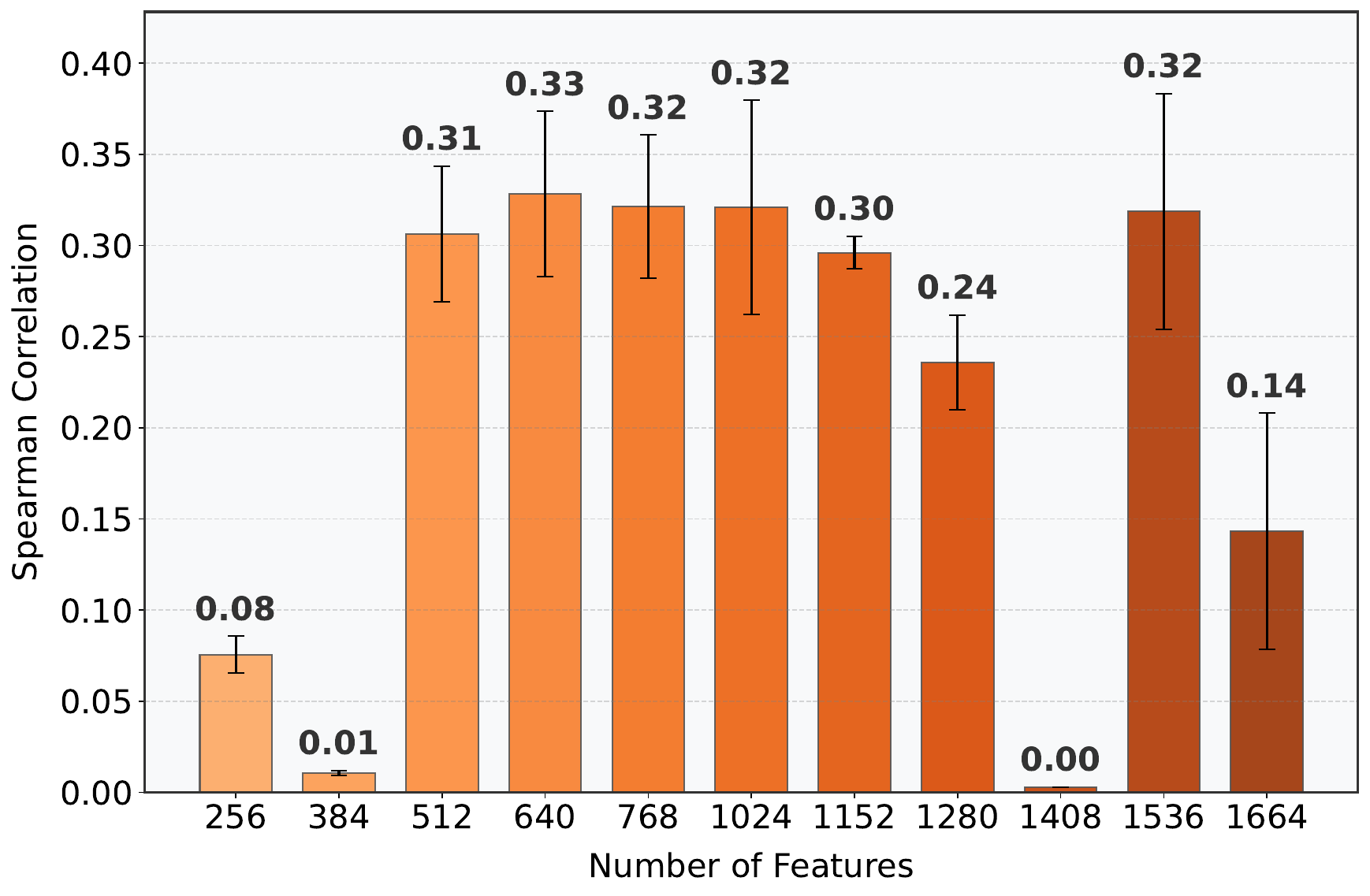}
            \caption{Effect of number of features.}
            \label{fig:COCO_subfig_num_features}
        \end{subfigure}\hfill
        \begin{subfigure}[b]{0.46\textwidth}
            \centering
            \includegraphics[width=\textwidth]{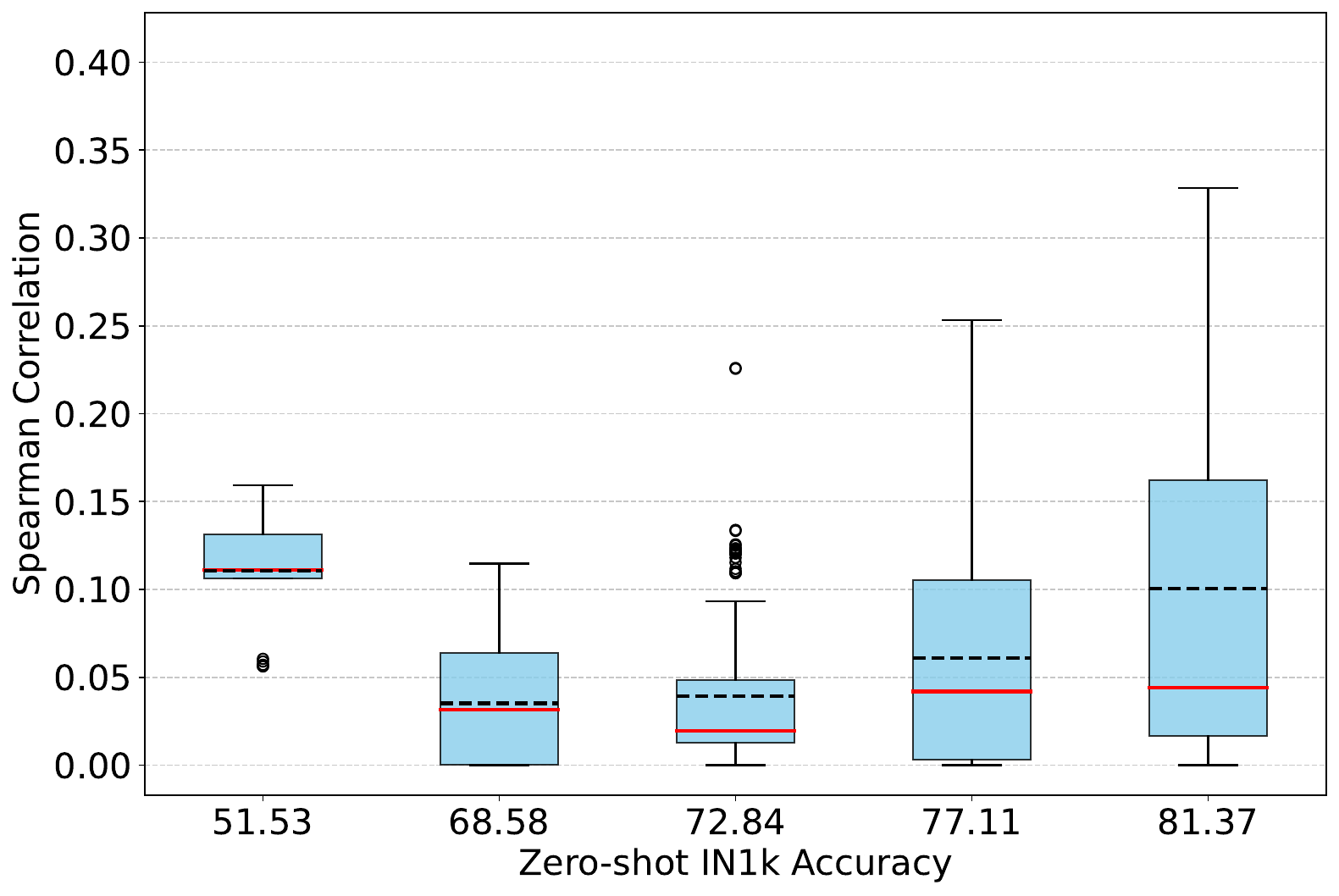}
            \caption{Effect of IN1K Zero-shot accuracy.}
            \label{fig:COCO_subfig_zs_in1k}
        \end{subfigure}
    \end{minipage}
    
    \caption{\textbf{Ablation study on randomly sampled COCO prompts} comparing the correlation to human preferences under varying factors: (a) the ViT training dataset, (b) input image size, (c) model capacity, (d) the number of features in the last ViT layer, and (e) zero-shot accuracy on ImageNet-1K.}
    \label{fig:COCO_ablation}
\end{figure*}

\noindent{\bf Effect of the size of the pre-training dataset on \metric.}
~\cref{fig:COCO_subfig_data_size} illustrates the correlation between \metric and human preferences as a function of the pre-training dataset size for Vision Transformer (ViT).
We observe a nonmonotonic relationship: increasing the pre-training dataset size does not consistently yield higher correlations. Notably, a model trained with fewer than 100 million samples achieves a high correlation of 0.33, while models trained with larger datasets show lower correlations ranging from 0.06 to 0.21. Interestingly, models trained with fewer than 5 billion samples demonstrate the lowest human correlation (0.06). These findings indicate that simply scaling up data does not guarantee improved performance for every metric or task, suggesting that balancing the quantity and quality of training data is crucial for optimal results.

\begin{figure*}[t]
    \centering
    % First row: three images
        \begin{subfigure}[b]{0.32\textwidth}
        \centering
        \includegraphics[width=\textwidth]{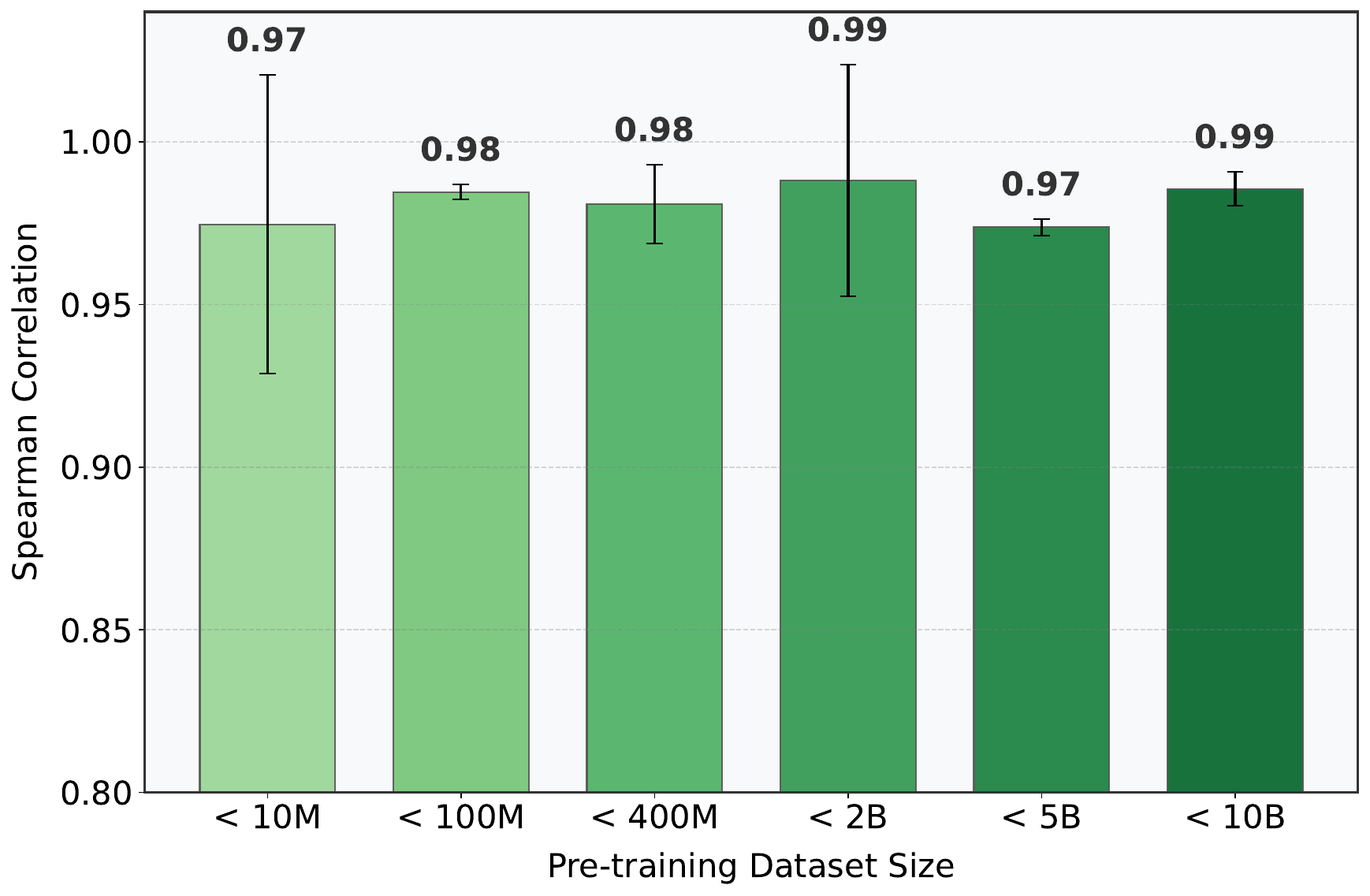}
        \caption{Effect of training data.}
        \label{fig:HPDv2_subfig_data_size}
\end{subfigure}
    \hfill
    \begin{subfigure}[b]{0.32\textwidth}
        \centering
        \includegraphics[width=\textwidth]{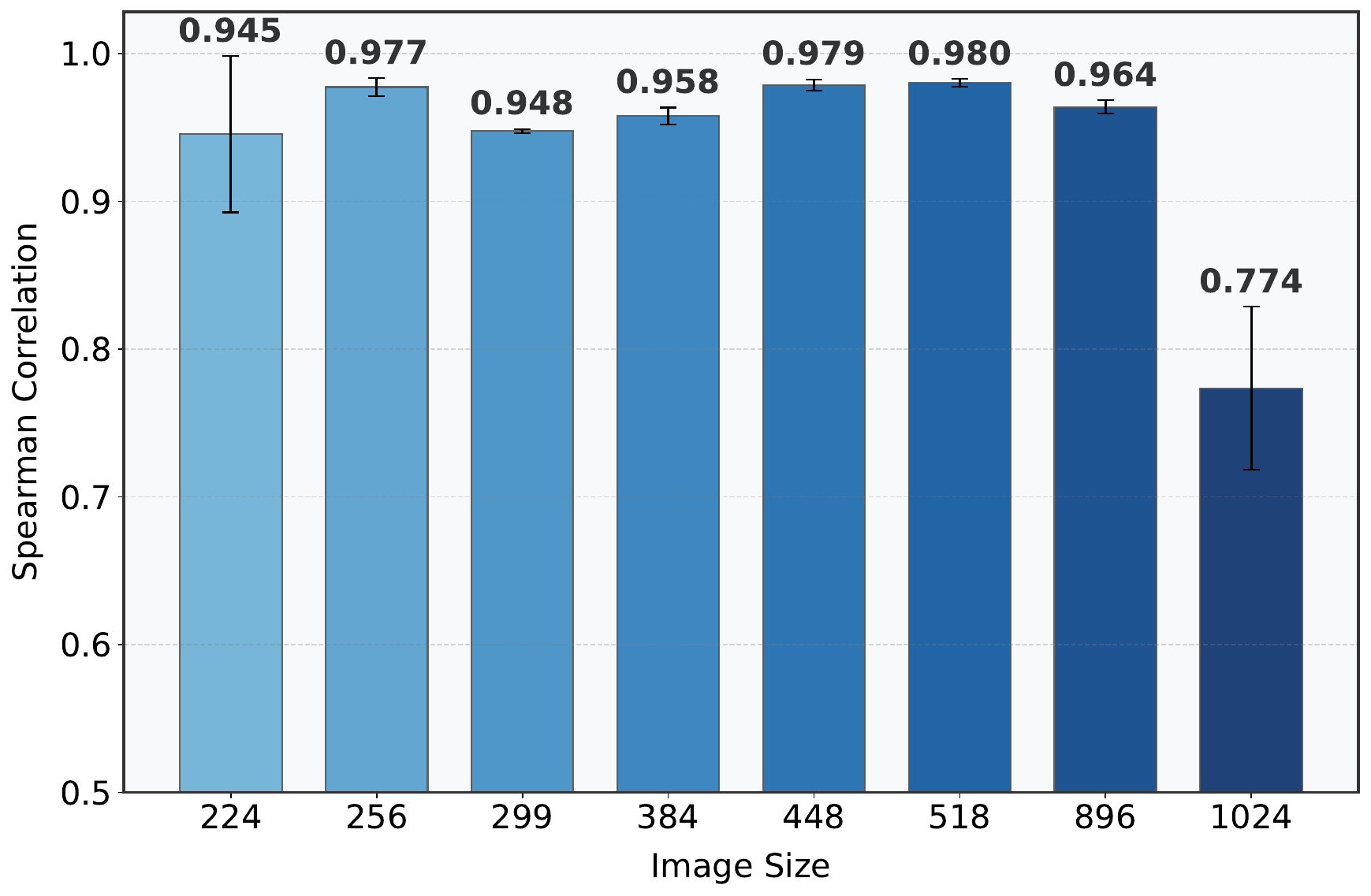}
        \caption{Effect of image size.}
        \label{fig:HPDv2_subfig_img_size}
    \end{subfigure}
    \hfill
    \begin{subfigure}[b]{0.32\textwidth}
        \centering
        \includegraphics[width=\textwidth]{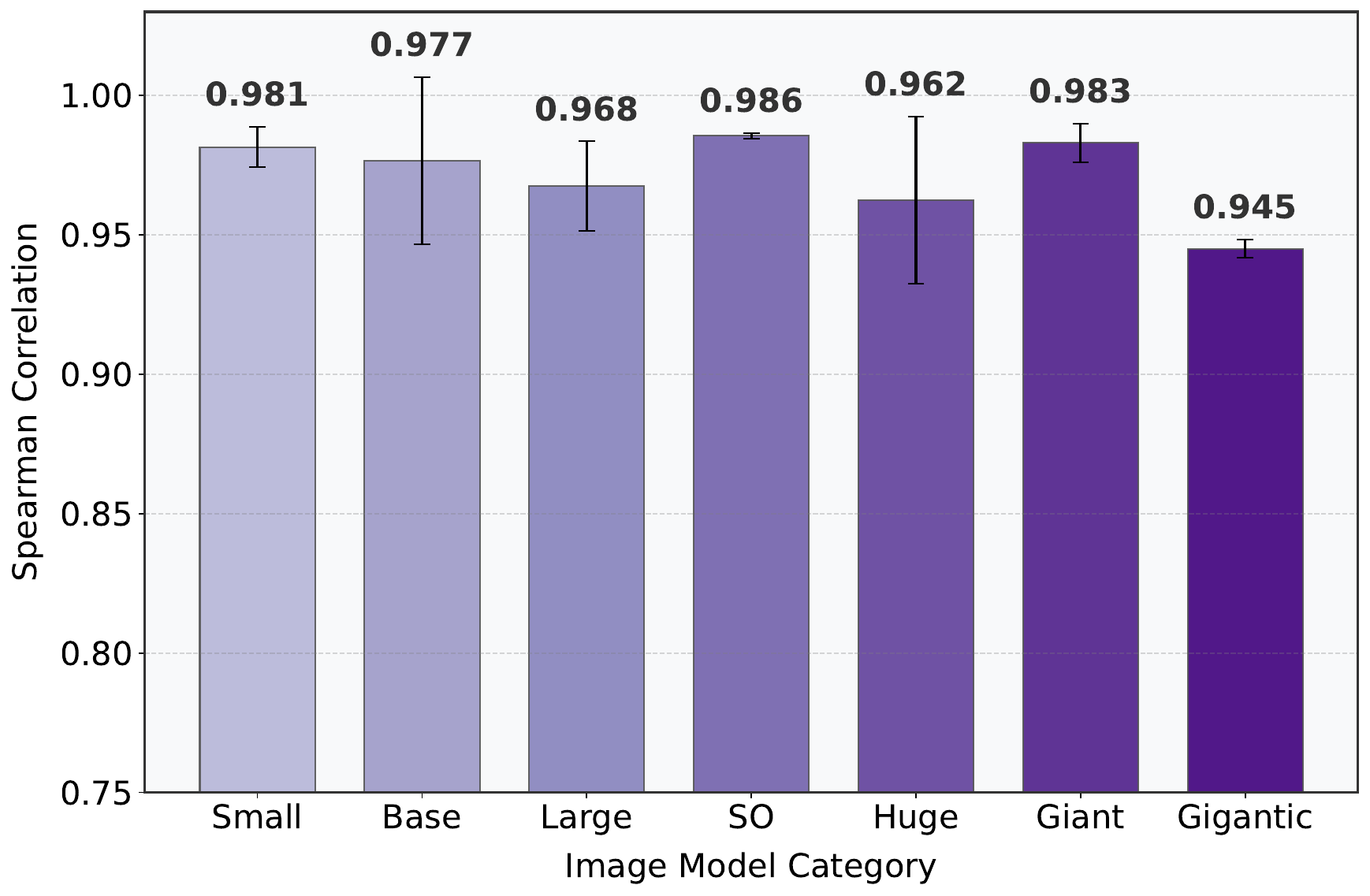}
         \caption{Effect of model size.}
        \label{fig:HPDv2_subfig_model_size}
    \end{subfigure}
    
    \vspace{1em}
    
    % Second row: two images centered under the three above
    \begin{minipage}{0.7\textwidth}
        \centering
        \begin{subfigure}[b]{0.46\textwidth}
            \centering
\includegraphics[width=\textwidth]{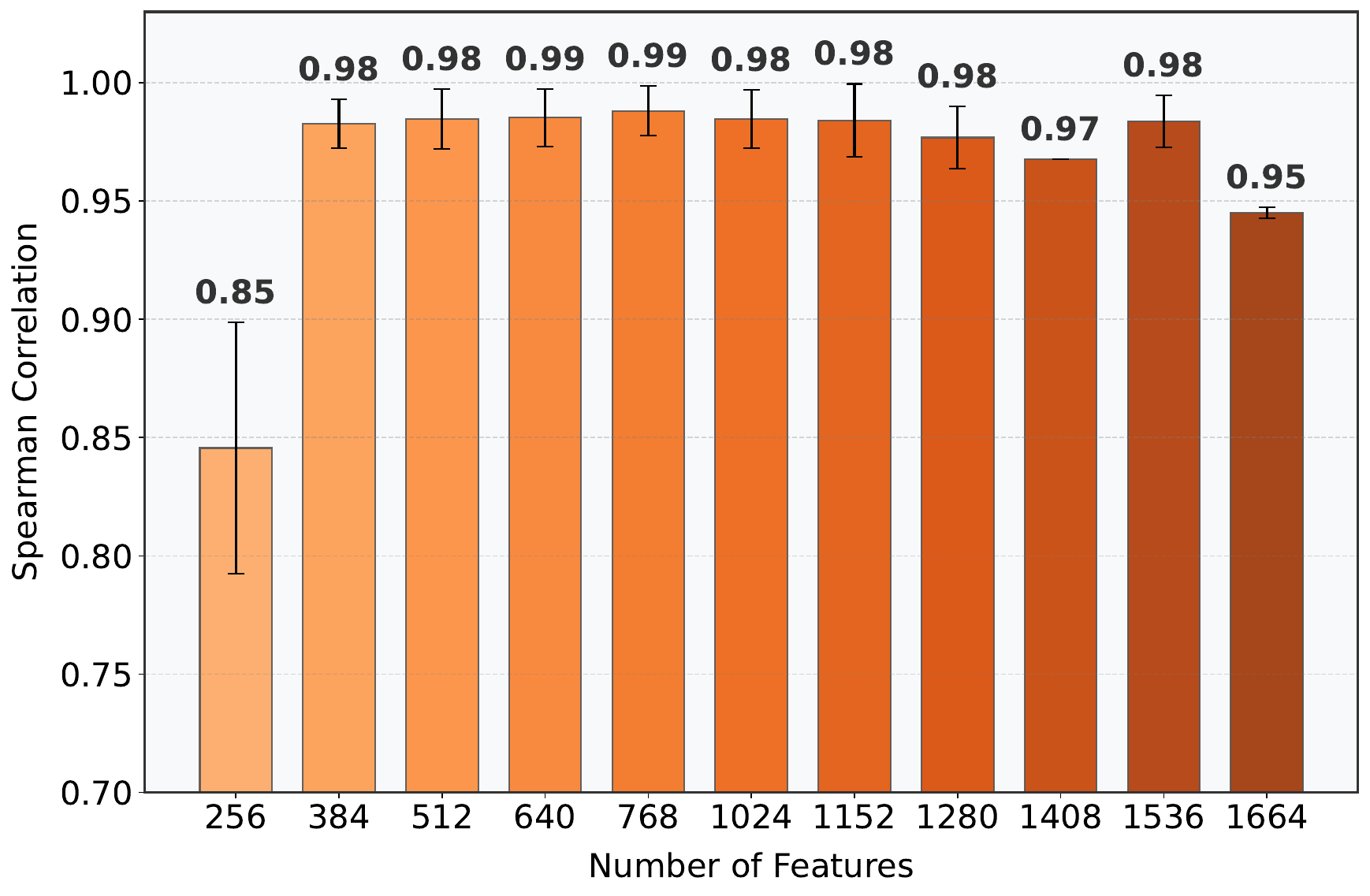}
            \caption{Effect of number of features.}
            \label{fig:HPDv2_subfig_num_features}
        \end{subfigure}\hfill
        \begin{subfigure}[b]{0.46\textwidth}
            \centering
            \includegraphics[width=\textwidth]{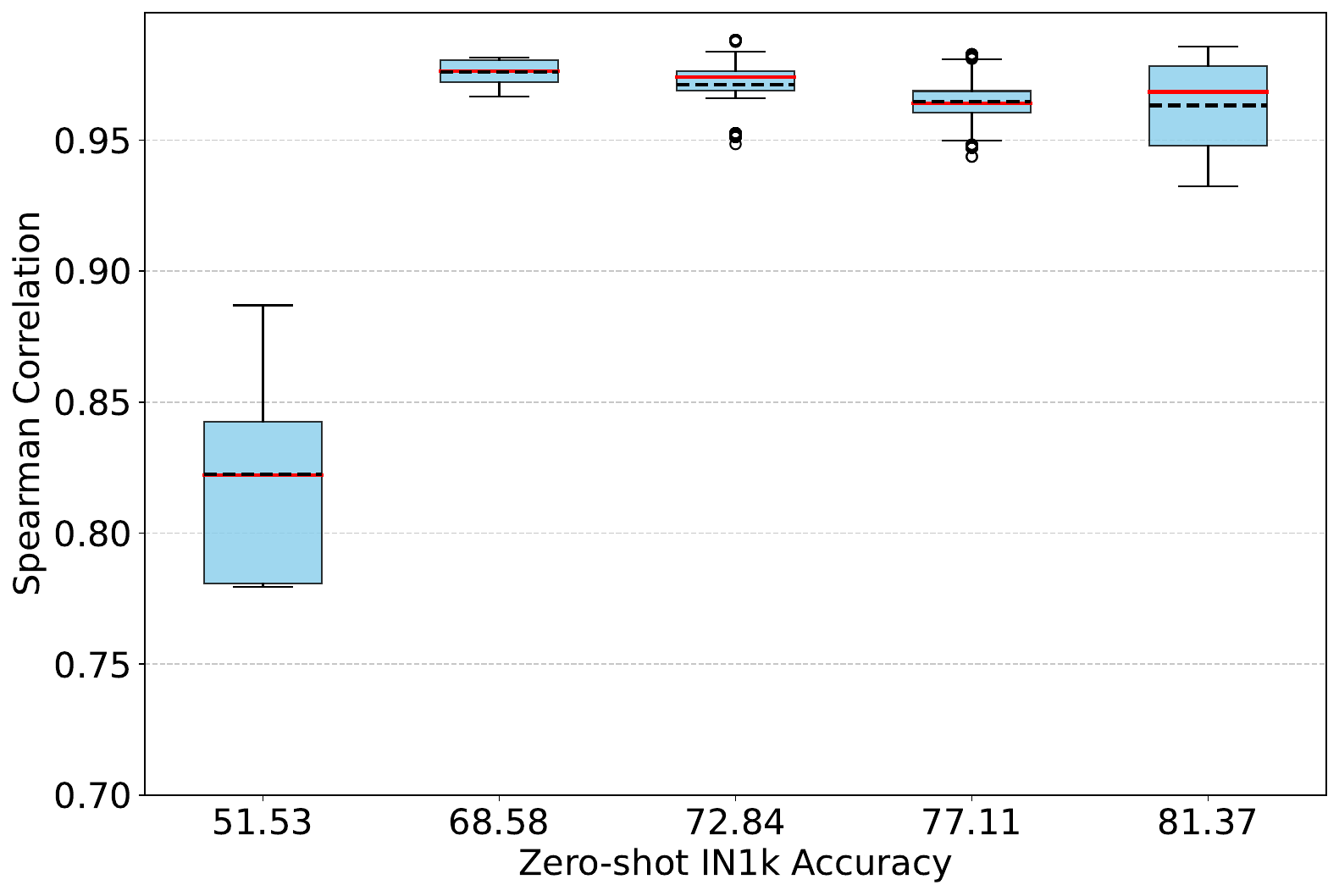}
            \caption{Effect of IN1K Zero-shot accuracy.}
            \label{fig:HPDv2_subfig_zs_in1k}
        \end{subfigure}
    \end{minipage}
    
    \caption{\textbf{Ablation study on the HPDv2} dataset comparing the correlation to human preferences under varying factors: (a) the ViT training dataset, (b) input image size, (c) model capacity, (d) the number of features in the last ViT layer, and (e) zero-shot accuracy on ImageNet-1K.}
    \label{fig:HPDv2_ablation}
\end{figure*}

\noindent{\bf Effect of image size on \metric.}
 ~\cref{fig:HPDv2_subfig_img_size} illustrates the correlation between \metric and human preferences across varying input image resolutions. We observe an inconsistent relationship: 
starting around 0.05 at an image size of 224x224, then rising to about 0.10 at 256x256 before dropping again to around 0.03 at 299x229. After a modest increase to 0.08 at 384x384, the correlation dips to 0.01 at 448x448 and then peaks sharply at 0.18 for an image size of 518x518. Beyond that, it decreases to nearly 0 at 896x896 and recovers slightly to 0.03 at 1024. 
These erratic trends suggest that there is no straightforward, monotonic relationship between image size and correlation for this particular task or metric, and the highest correlation appears in the midrange rather than at the smallest or largest resolutions.

\noindent{\bf Effect of ViT model size on \metric.}
The correlation between \metric and human preferences across Vision Transformer (ViT) sizes are presented in~\cref{fig:HPDv2_subfig_model_size}. Larger models tend to improve alignment with human judgments, with \emph{Giant} model achieving the highest correlation of 0.320. However, when the model size gets bigger to \emph{Gigantic} model, the correlation degrades down to 0.089.
These results indicate that correlation does not simply increase in tandem with model size; rather, there seems to be an optimal range, as exemplified by the \emph{Giant} model, for achieving the strongest alignment with the evaluation metric.

\noindent{\bf Effect of ViT Feature dimensionality on \metric.}
 ~\cref{fig:HPDv2_subfig_num_features} shows the correlation between \metric and human preferences across ViT feature dimensions from 256 to 1664. The two lowest numbers of the features (256 and 384) show the lowest correlation. However, we observe a clear plateau effect in performance once the feature count reaches 512, with correlation values stabilizing around 0.30-0.32 across a wide range of dimensionalities (512-1152). Interestingly, at extremely high dimensionalities (above 1536), we note a slight performance decline, with correlation dropping to 0 at the 1408 feature. Although the correlation increases back to 0.32 with 1536 features, it decreases back to 0.14 at 1664 features. This suggests an optimal range for feature dimensionality exists, beyond which additional computational complexity yields diminishing or even negative returns.

\noindent{\bf Effect of Zero-Shot ImageNet Accuracy on \metric.}
 ~\cref{fig:HPDv2_subfig_zs_in1k} depicts a boxplot of the correlation between \metric and human preferences as a function of zero-shot ImageNet accuracy, evaluated exclusively on image-text pretrained models~\cite{radford2021learning}. Interestingly, we find that the lowest zero-shot accuracy has the highest correlation to human preference.  However, after 68.58 of zero-shot accuracies, higher zero-shot accuracies generally correspond to stronger correlations with human judgments, though variance exists within each accuracy bin.

\begin{table*}[ht]
    \centering
    \small
    \begin{tabular}{|l|l|}
        \hline
        \textbf{Model Name} & \textbf{Model Path} \\
        \hline
        ViT-S/16 (IN1K) & \texttt{timm/vit\_small\_patch16\_224.augreg\_in1k} \\
        ViT-B/16 (IN1K) & \texttt{timm/vit\_base\_patch16\_224.augreg\_in1k} \\
        ViT-S/16 (IN21K) & \texttt{timm/vit\_small\_patch16\_224.augreg\_in21k} \\
        ViT-B/16 (IN21K) & \texttt{timm/vit\_base\_patch16\_224.augreg\_in21k} \\
        ViT-L/16 (IN21K) & \texttt{timm/vit\_large\_patch16\_224.augreg\_in21k} \\
        ViT-H/14 (IN21K) & \texttt{timm/vit\_huge\_patch14\_224.orig\_in21k} \\
        MAE-B/16 & \texttt{timm/vit\_base\_patch16\_224.mae} \\
        MAE-L/16 & \texttt{timm/vit\_large\_patch16\_224.mae} \\
        MAE-H/14 & \texttt{timm/vit\_huge\_patch14\_224.mae} \\
        DINOv2-S/14 & \texttt{timm/vit\_small\_patch14\_dinov2.lvd142m} \\
        DINOv2-B/14 & \texttt{timm/vit\_base\_patch14\_reg4\_dinov2.lvd142m} \\
        DINOv2-L/14 & \texttt{timm/vit\_large\_patch14\_dinov2.lvd142m} \\
        DINOv2-G/14 & \texttt{timm/vit\_giant\_patch14\_dinov2.lvd142m} \\
        SAM-ViT-B/16 & \texttt{timm/samvit\_base\_patch16} \\
        SAM-ViT-H/16 & \texttt{timm/samvit\_huge\_patch16} \\
        MoCov3-ViT-B & \texttt{nyu-visionx/moco-v3-vit-b} \\
        MoCov3-ViT-L & \texttt{nyu-visionx/moco-v3-vit-l} \\
        I-JEPA-H/14 (IN1K) & \texttt{jmtzt/ijepa\_vith14\_1k} \\
        I-JEPA-H/14 (IN21K) & \texttt{facebook/ijepa\_vith14\_22k} \\
        CLIP-B/16 & \texttt{timm/vit\_base\_patch16\_clip\_224.openai} \\
        CLIP-L/14 & \texttt{timm/vit\_large\_patch14\_clip\_224.openai} \\
        MetaCLIP-B/16 & \texttt{timm/vit\_base\_patch16\_clip\_224.metaclip\_400m} \\
        MetaCLIP-L/14 & \texttt{timm/vit\_large\_patch14\_clip\_224.metaclip\_400m} \\
        MetaCLIP-H/14 & \texttt{timm/vit\_huge\_patch14\_clip\_224.metaclip\_altogether} \\
        MetaCLIP-G/14 & \texttt{timm/vit\_gigantic\_patch14\_clip\_224.metaclip\_2pt5b} \\
        DFN-CLIP-B/16 & \texttt{timm/vit\_base\_patch16\_clip\_224.dfn2b} \\
        DFN-CLIP-L/14 & \texttt{timm/vit\_large\_patch14\_clip\_224.dfn2b} \\
        DFN-CLIP-H/14 & \texttt{timm/vit\_huge\_patch14\_clip\_224.dfn5b} \\
        OpenCLIP-B/16 & \texttt{timm/vit\_base\_patch16\_clip\_224.laion2b} \\
        OpenCLIP-L/14 & \texttt{timm/vit\_large\_patch14\_clip\_224.laion2b} \\
        OpenCLIP-H/14 & \texttt{timm/vit\_huge\_patch14\_clip\_224.laion2b} \\
        OpenCLIP-g/14 & \texttt{timm/vit\_giant\_patch14\_clip\_224.laion2b} \\
        OpenCLIP-G/14 & \texttt{timm/vit\_gigantic\_patch14\_clip\_224.laion2b} \\
        DataComp-CLIP-B/16 & \texttt{timm/vit\_base\_patch16\_clip\_224.datacompxl} \\
        DataComp-CLIP-L/14 & \texttt{timm/vit\_large\_patch14\_clip\_224.datacompxl} \\
        EVA02-B/16 & \texttt{timm/eva02\_base\_patch16\_clip\_224} \\
        EVA02-L/14 & \texttt{timm/eva02\_large\_patch14\_clip\_224} \\
        ConvNeXT-CLIP-B & \texttt{timm/convnext\_base.clip\_laion2b} \\
        ConvNeXT-CLIP-B(A) & \texttt{timm/convnext\_base.clip\_laiona} \\
        ConvNeXT-CLIP-L & \texttt{timm/convnext\_large\_mlp.clip\_laion2b\_augreg} \\
        SigLIP-B/16 & \texttt{timm/vit\_base\_patch16\_siglip\_224.webli} \\
        SigLIP-L/16 & \texttt{timm/vit\_large\_patch16\_siglip\_256.webli} \\
        SigLIP-SO/14 & \texttt{timm/vit\_so400m\_patch14\_siglip\_gap\_224.webli} \\
        InceptionV3 & \texttt{inception/inceptionv3} \\
        SigLIP-SO/14 (448) & \texttt{timm/vit\_so400m\_patch14\_siglip\_gap\_448.pali2\_10b\_pt} \\
        SigLIP-SO/14 (896) & \texttt{timm/vit\_so400m\_patch14\_siglip\_gap\_896.pali2\_10b\_pt} \\
        \hline
    \end{tabular}
    \caption{List of image backbone models used in our analysis.}
    \label{tab:image_model_list}
\end{table*}
\begin{table*}[ht]
    \centering
    \small
    \begin{tabular}{|l|l|}
        \hline
        \textbf{Model Name} & \textbf{Model Path} \\
        \hline
        RoBERTa-base & \texttt{FacebookAI/roberta-base} \\
        RoBERTa-large & \texttt{FacebookAI/roberta-large} \\
        XLM-RoBERTa-base & \texttt{FacebookAI/xlm-roberta-base} \\
        XLM-RoBERTa-large & \texttt{FacebookAI/xlm-roberta-large} \\
        BERT-base & \texttt{google-bert/bert-base-uncased} \\
        BERT-large & \texttt{google-bert/bert-large-uncased} \\
        ModernBERT-base & \texttt{answerdotai/ModernBERT-base} \\
        ModernBERT-large & \texttt{answerdotai/ModernBERT-large} \\
        ALBERT-base-v2 & \texttt{albert/albert-base-v2} \\
        ALBERT-large-v2 & \texttt{albert/albert-large-v2} \\
        ALBERT-xlarge-v2 & \texttt{albert/albert-xlarge-v2} \\
        ALBERT-xxlarge-v2 & \texttt{albert/albert-xxlarge-v2} \\
        FLAN-T5-B & \texttt{google/flan-t5-base} \\
        FLAN-T5-L & \texttt{google/flan-t5-large} \\
        FLAN-T5-XL & \texttt{google/flan-t5-xl} \\
        FLAN-T5-XXL & \texttt{google/flan-t5-xxl} \\
        T5-base & \texttt{google/t5-v1\_1-base} \\
        T5-large & \texttt{google/t5-v1\_1-large} \\
        T5-xlarge & \texttt{google/t5-v1\_1-xl} \\
        T5-xxlarge & \texttt{google/t5-v1\_1-xxl} \\
        CLIP-B/16 & \texttt{ViT-B-16-quickgelu.openai} \\
        CLIP-L/14 & \texttt{ViT-L-14-quickgelu.openai} \\
        MetaCLIP-B/16 & \texttt{ViT-B-16-quickgelu.metaclip\_400m} \\
        MetaCLIP-L/14 & \texttt{ViT-L-14-quickgelu.metaclip\_400m} \\
        MetaCLIP-H/14 & \texttt{ViT-H-14.metaclip\_altogether} \\
        MetaCLIP-G/14 & \texttt{ViT-bigG-14-CLIPA.datacomp1b} \\
        DFN-CLIP-B/16 & \texttt{ViT-B-16-quickgelu.dfn2b} \\
        DFN-CLIP-L/14 & \texttt{ViT-L-14-quickgelu.dfn2b} \\
        DFN-CLIP-H/14 & \texttt{ViT-H-14-quickgelu.dfn5b} \\
        OpenCLIP-B/16 & \texttt{ViT-B-16.laion2b\_s34b\_b88k} \\
        OpenCLIP-L/14 & \texttt{ViT-L-14.laion2b\_s32b\_b82k} \\
        OpenCLIP-H/14 & \texttt{ViT-H-14.laion2b\_s32b\_b79k} \\
        OpenCLIP-g/14 & \texttt{ViT-g-14.laion2b\_s34b\_b88k} \\
        DataComp-CLIP-B/16 & \texttt{ViT-B-16.datacomp\_xl\_s13b\_b90k} \\
        DataComp-CLIP-L/14 & \texttt{ViT-L-14.datacomp\_xl\_s13b\_b90k} \\
        EVA02-B/16 & \texttt{EVA02-B-16.merged2b\_s8b\_b131k} \\
        EVA02-L/14 & \texttt{EVA02-L-14.merged2b\_s4b\_b131k} \\
        ConvNeXT-CLIP-B & \texttt{convnext\_base\_w.laion2b\_s13b\_b82k\_augreg} \\
        ConvNeXT-CLIP-B(A) & \texttt{convnext\_base\_w.laion\_aesthetic\_s13b\_b82k} \\
        ConvNeXT-CLIP-L & \texttt{convnext\_large\_d.laion2b\_s26b\_b102k\_augreg} \\
        SigLIP-B/16 & \texttt{ViT-B-16-SigLIP.webli} \\
        SigLIP-L/16 & \texttt{ViT-L-16-SigLIP-256.webli} \\
        SigLIP-SO/14 & \texttt{ViT-SO400M-14-SigLIP.webli} \\
        \hline
    \end{tabular}
    \caption{List of text backbone models used in our analysis.}
    \label{tab:text_model_list}
\end{table*}

\end{document}